\documentclass{article} 
\usepackage{iclr2026_conference,times}

\usepackage{hyperref}
\usepackage{url}

\usepackage{booktabs}       
\usepackage{amsfonts}       
\usepackage{nicefrac}       
\usepackage{microtype}      
\usepackage{xcolor}         
\usepackage{enumitem} 
\usepackage{amsmath}
\usepackage{amssymb}
\usepackage{mathtools}
\usepackage{amsthm}
\usepackage[capitalise]{cleveref}

\usepackage{algorithm}
\usepackage{algpseudocode}

\usepackage{listings}
\usepackage{xcolor}  
\definecolor{vscbg}{HTML}{1E1E1E}      
\definecolor{vscblue}{HTML}{569CD6}    
\definecolor{vscgreen}{HTML}{6A9955}   
\definecolor{vscstring}{HTML}{CE9178}  
\definecolor{vsctext}{HTML}{D4D4D4}    

\lstdefinestyle{vscode}{
  backgroundcolor=\color{vscbg},
  basicstyle=\ttfamily\color{vsctext}\footnotesize,
  keywordstyle=\color{vscblue}\bfseries,
  commentstyle=\color{vscgreen}\itshape,
  stringstyle=\color{vscstring},
  showstringspaces=false,
  breaklines=true,
  tabsize=4,
  frame=single,
  numbers=left,
  numberstyle=\tiny\color{gray},
  captionpos=b,
  language=Python
}

\newcommand{\model}{MP-SSM}
\newcommand{\X}{\mathbf{X}} 
\newcommand{\W}{\mathbf{W}} 
\newcommand{\B}{\mathbf{B}} 
\newcommand{\A}{\mathbf{A}} 

\newcommand{\D}{\mathbf{D}^{-\frac{1}{2}}}
\newcommand{\q}{\mathbf{q}}

\newcommand{\U}{\mathbf{U}}

\newcommand{\M}{\mathbf{M}}
\newcommand{\Y}{\mathbf{Y}}
\newcommand{\I}{\mathbf{I}}

\newcommand{\Z}{\mathbf{Z}}
\newcommand{\V}{\mathbf{V}}

\newcommand{\x}{\mathbf{x}}
\newcommand{\y}{\mathbf{y}}
\newcommand{\Q}{\mathbf{Q}}
\newcommand{\Lm}{\mathbf{\Lambda}}
\newcommand{\PP}{\mathbf{P}}

\newcommand{\ourmethod}{MP-SSM}
\newcommand{\ie}{i.e., }

\newcommand{\revise}[1]{\textcolor{red}{#1}} 
\definecolor{first}{HTML}{00A64F}  
\definecolor{second}{HTML}{006EB8} 
\newcommand{\one}[1]{\textcolor{first}{\bf#1}}
\newcommand{\two}[1]{\textcolor{second}{\bf#1}}
\newcommand{\three}[1]{{\bf#1}}

\usepackage{multirow}
\usepackage{wrapfig}

\theoremstyle{plain}
\newtheorem{theorem}{Theorem}[section]

\newtheorem{lemma}[theorem]{Lemma}
\newtheorem{corollary}[theorem]{Corollary}
\theoremstyle{definition}
\newtheorem{definition}[theorem]{Definition}

\theoremstyle{remark}
\newtheorem{remark}[theorem]{Remark}

\title{Message-Passing State-Space Models: Improving Graph Learning with Modern Sequence Modeling 
}

\author{Andrea Ceni\\ 
University of Pisa \\
\texttt{andrea.ceni@unipi.it}
\And
Alessio Gravina\\
University of Pisa \\
\texttt{alessio.gravina@di.unipi.it}
\And
Claudio Gallicchio \\
University of Pisa \\
\texttt{claudio.gallicchio@unipi.it}
\And
Davide Bacciu \\
University of Pisa \\
\texttt{davide.bacciu@unipi.it}
\And
Carola-Bibiane Sch\"onlieb \\
University of Cambridge\\
\texttt{cbs31@cam.ac.uk}
\And
Moshe Eliasof \\
University of Cambridge\\
\texttt{me532@cam.ac.uk}
}

\iclrfinalcopy 
\begin{document}

\maketitle

\begin{abstract}
The recent success of State-Space Models (SSMs) in sequence modeling has inspired their extension to graphs, giving rise to Graph State-Space Models (GSSMs). While effective, existing approaches often rely on sequentializations or spectral decompositions that lack permutation equivariance, message-passing compatibility, and computational efficiency. Moreover, they typically target either static or temporal graphs in isolation and, crucially, provide only loose or qualitative results on information propagation, offering 
no exact guarantees 
%
on challenges such as vanishing gradients and over-squashing. In this work, we revisit the design of GSSMs through the lens of sensitivity analysis. We introduce a principled integration of modern SSM computation into the Message-Passing Neural Network framework, yielding a unified architecture that is computationally efficient, permutation equivariant, and supports fast parallelism. Our formulation admits closed-form Jacobian computations, enabling an exact sensitivity analysis of node-to-node dependencies and rigorous lower bounds on information flow, contrasting sharply with prior heuristic approaches. These theoretical insights clarify when and how stable long-range propagation can be achieved. Finally, we validate our model across a wide range of benchmarks, including node classification, graph property prediction, long-range reasoning, and spatiotemporal forecasting, where it achieves strong empirical performance while preserving the simplicity of message passing.
\end{abstract}


\section{Introduction} 
\label{sec:intro}

Graph Neural Networks (GNNs) \citep{GNNsurvey,gravina2023survey}, and in particular Message-Passing Neural Networks (MPNNs) \citep{MPNN}, have become a standard tool for learning from graph-structured data. Yet, traditional MPNNs such as GCNs \citep{Kipf2016} struggle to propagate information across distant nodes due to phenomena like over-squashing \citep{alon2021on, topping2022understanding,di2023over} and, more generally, vanishing gradients \citep{di2023over, pmlr-v28-pascanu13, arroyo25vanishing}, which limit their effectiveness on tasks requiring long-range dependency modeling \citep{dwivedi2022long}. While a variety of remedies have been proposed, ranging from rewiring techniques \citep{topping2022understanding, karhadkar2023fosr,drew}, to transformers \citep{kreuzer2021rethinking,ying2021transformers,graphgps,Dwivedi2021,dwivedi2022graph}, to regularization strategies in weight space \citep{gravina_adgn, gravina_swan}, these often rely on heavy architectural modifications and typically do not extend cleanly to standard MPNNs like GCNs.

In parallel, State-Space Models (SSMs) have recently emerged as a powerful paradigm in sequence modeling, with architectures such as LRU \citep{orvieto2023resurrecting}, S4 \citep{gu2021efficiently}, and subsequent extensions \citep{smith2022simplified,gupta2022diagonal,poli2023hyena,fu2022hungry}, culminating in advanced designs like Mamba \citep{mamba}, Griffin \citep{de2024griffin}, and xLSTM \citep{beck2024xlstm}. These models rely on linear recurrent dynamics interleaved with nonlinear projections, a design that enables efficient training, stable gradient flow, universal approximation, and robust long-range dependency modeling \citep{pipiras2017long,voelker2019legendre,orvieto2024universality,muca2024theoretical}. Inspired by this progress, several works have attempted to adapt SSMs to graph learning. Current approaches, however, either enforce sequentializations of the graph \citep{tang2023modeling,wang2024mamba,behrouz2024graph} or adopt spectral decompositions \citep{huang2024can,zhao2024grassnet}, which may compromise permutation equivariance \citep{bronstein2021geometric}, distort graph topology, or rely on non-unique modes \citep{lim2023sign}. Moreover, while these methods improve propagation in practice, they provide at best loose  guarantees on sensitivity, leaving fundamental questions about stability and information flow unanswered.

This paper revisits Graph State-Space Models (GSSMs) through the lens of sensitivity analysis, \ie by studying how the state of a node depends on information from distant nodes. We propose a principled integration of SSM computation into the MPNN framework that not only preserves permutation equivariance and computational efficiency, but also admits \emph{exact} sensitivity analysis. This allows us to rigorously quantify node-to-node dependencies, derive precise lower bounds for vanishing gradients and over-squashing, and identify unfavorable topologies that exacerbate propagation bottlenecks. In contrast to prior work that relies on approximations or heuristic arguments, our analysis provides concrete and informative characterizations of how information flows in the deep regime.

Our contributions can be summarized as follows:
\begin{enumerate}
    \item 
    \textbf{Principled integration of SSMs and MPNNs through sensitivity analisys.} We introduce a simple yet general framework, namely Message-Passing State-Space Model (MP-SSM), that embeds linear state-space dynamics directly into message passing. 
    This design 
    unifies static and temporal graphs while preserving permutation equivariance and graph topology. It enables stable long-range information propagation and supports fast parallel implementation.
    \item 
    \textbf{Exact sensitivity analysis.} Our formulation enables closed-form Jacobian computations, yielding exact characterizations of local and global sensitivities, \ie the model’s information transfer capacity. We provide lower bounds that directly link architectural design choices to the alleviation of vanishing gradients and over-squashing.
    \item 
    \textbf{Empirical validation.} Across 15 benchmarks including synthetic and real-world long-range tasks, heterophilic node classification, and spatiotemporal forecasting, our approach consistently matches or outperforms state-of-the-art baselines, demonstrating both its versatility and effectiveness.
\end{enumerate}

\section{Message-Passing State-Space Model}\label{sec:model}
In this section, we present our framework, called \textit{Message-Passing State-Space Model} (\ourmethod), which integrates state-space modeling into the message-passing paradigm. The theoretical analysis that guided the design of \ourmethod{} and ensures principled information propagation across the graph is detailed in \cref{sec:sensitivity}.


\begin{figure}
    \centering
    \includegraphics[width=0.9\linewidth]{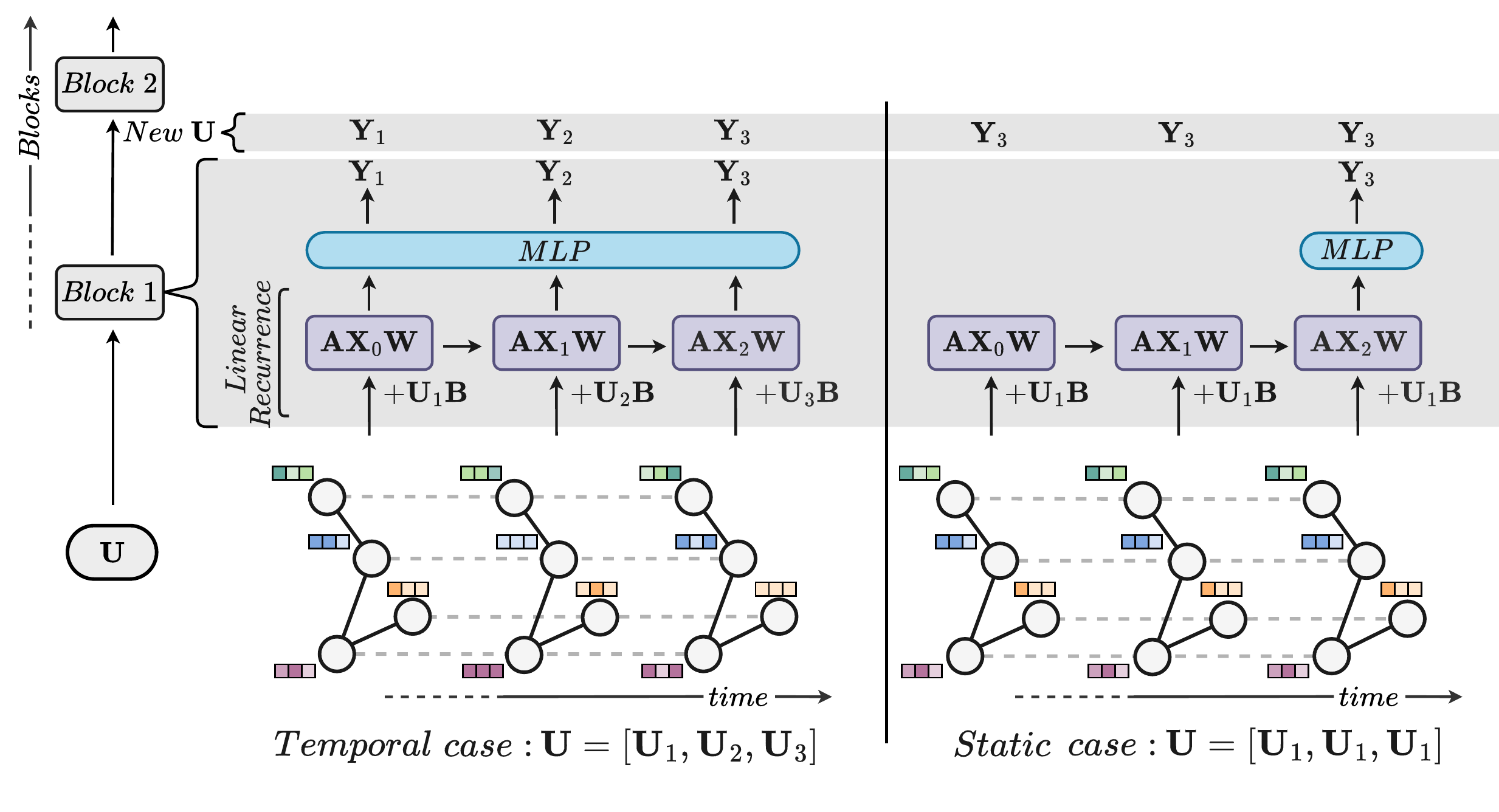}
    \vspace{-2mm}
    \caption{Illustration of our \model{} for temporal and static cases, considering a recurrence time $k+1=3$. The temporal case (left) incorporates dynamic updates to node embeddings over time steps, represented as $\U=[\U_1, \U_2, \U_3]$, while the static case (right) uses fixed node embeddings $\U=[\U_1, \U_1, \U_1]$. An \model{} block comprises a linear recurrence followed by a multilayer perceptron (MLP). Multiple \model{} blocks are stacked to construct a deep \ourmethod{} architecture. 
    }
    \label{fig:model}
\end{figure}

\textbf{Graph and Shift Operator.} We represent a graph as ${G=(V,E)}$ where ${V}$ is a set of
$n$ nodes, and ${E} \subseteq {V} \times {V}$ is a set of $m$ undirected edges. The adjacency matrix $\tilde{\mathbf{A}} \in \{0,1\}^{n \times n}$ encodes edge presence with $ (\tilde{\mathbf{A}})_{ij}=1 $ if $(i,j) \in {E}$, and zero otherwise. 
To enable message passing, we use the graph shift operator (GSO) defined as the symmetrically normalized adjacency with self-loops \citep{Kipf2016}:
\begin{equation}
    \label{eq:shift_operator}
    {\A} = \D (\tilde{\A}+ \I) \D,
\end{equation}
where $\tilde{\A}$ is the adjacency matrix, and $\mathbf{D}$ is the degree matrix of $\tilde{\A} + \I$, with $(\D)_{ii} = \bigl(1+ \sum_{j=1}^{n} (\tilde{\A})_{ij}\bigl)^{-\frac{1}{2}}$.
We emphasize that, although we adopt the GSO in \cref{eq:shift_operator} for its simplicity and widespread use, our framework is compatible with any choice of GSO.

\textbf{Linear State-Space Recurrence on Graphs.}
\label{sec:state_space_recurrence}
We denote the graph data as a sequence of input node features $ [ \U_t ]_{t=1}^{T}$, with $\U_t \in \mathbb{R}^{n \times c' }$, with $c'$ being the dimensionality of the input features. We note that, for static graphs the sequence consists of a single element, \ie $\U_1$ (as shown in the bottom-right of \cref{fig:model}).
We embed the sequence of input states, obtaining a sequence of hidden states $ [ \X_t ]_{t=1}^{T}$, where $ \X_t \in \mathbb{R}^{n \times c}$, via a linear message-passing scheme and channel mixing with learnable weight matrices $\W \in \mathbb{R}^{c \times c},\B \in \mathbb{R}^{c' \times c}$, as follows:
\begin{equation}
\label{eq:graph_dynamics}
    \X_{t+1} = \A \X_{t} \W +  \U_{t+1}\B  .
\end{equation}
\cref{eq:graph_dynamics} represents the linear state-space recurrence on graphs. 
Note that the message-passing mechanism of many popular GNN models in the literature can be expressed through the form of 
of this equation, including  methods like GCN \citep{Kipf2016}, 
GAT \citep{velivckovic2017graph}, and GIN \citep{xu2018how}.
\emph{A key distinction from such models lies in the use of a purely linear recurrent equation.}
This design choice is consistent with modern SSM approaches and, crucially, enables both an exact sensitivity analysis (discussed in \cref{sec:sensitivity}) and an efficient parallel implementation. Specifically, \ourmethod{} can be parallelised by unrolling the linear recurrence and computing a closed-form solution in a single step. In Appendix~\ref{app:fast_implementation}, we describe our fast implementation, discussing both its advantages and limitations, and provide a runtime comparison with a standard GCN, showing that \ourmethod{} can achieve up to a 1000× speedup.

\textbf{\model{} Block.} A block of our \ourmethod{} is designed to propagate information between nodes that are $k$ hops away from each other, where  $k$ can also be large, as discussed in \cref{sec:sensitivity}. Each block is composed of $k$ iterations of the linear recurrence described in \cref{eq:graph_dynamics}, followed by a learnable graph-agnostic nonlinear mapping.
Setting the initial state $\X_0 = \mathbf{0}\in \mathbb{R}^{n\times c}$, 
we define our \ourmethod{} block as:
\begin{align}
\label{eq:state_space}
    \X_{t+1} &= \A \X_{t} \W +  \U_{t+1}\B , \qquad t=0,\ldots,k,  \\
\label{eq:mlp}
    \Y_{t+1} &=  \text{MLP}(\X_{t+1}),
\end{align}
where MLP denotes a nonlinear multilayer perceptron of $2$ dense layers with a nonlinearity in between them, and $k$ a hyperparameter defining the depth of each \ourmethod{} block.
\cref{eq:state_space,eq:mlp} define the state-space representation on graphs, which forms the foundation of our proposed \model{}.
Our framework is inspired by SSMs, which are naturally suited for sequential data. 
In temporal graph settings, the input naturally consists of a sequence of graphs (e.g., with time-varying features). Given an input sequence $\U = [\U_1, \U_2, \dots, \U_{k+1}]$, we apply the same MLP decoder of \cref{eq:mlp}, shared across all time steps, to the corresponding embeddings $[\X_1, \X_2, \dots, \X_{k+1}]$, producing an output sequence $[\Y_1, \Y_2, \dots, \Y_{k+1}]$ of the same length.
For static graphs, however, we must construct a sequence from a single input instance $\U_1$. 
As illustrated in Figure~\ref{fig:model}, we unify the treatment of temporal and static settings by generating a constant input sequence $\U = [\U_1, \U_1, \ldots, \U_1]$ of length $k+1$ for the static case. We note that this design induces a skip connection in the recurrence. 
In the static setting, the MLP decoder is applied solely to the final embedding after $k+1$ steps. Consequently, both the input and output sequences are constant: $[\U_1, \U_1, \dots, \U_1]$ as input and $[\Y_{k+1}, \Y_{k+1}, \dots, \Y_{k+1}]$ as output.
\cref{fig:model} illustrates and summarizes the modes of operation described above.
%
%
In \cref{app:other_temporal_models}, we clarify the originality of our framework in relation to recent 
state-space modeling approaches for temporal graphs, like GGRNN \citep{ruiz2020gated} and GraphSSM \citep{li2024state}. A key feature that distinguishes \ourmethod{} from these approaches is the absence of nonlinearity in the graph diffusion dynamics. In fact, the only nonlinearity in the entire \ourmethod{} block resides within the MLP decoder. This property is crucial both for enabling exact sensitivity analysis and for supporting an efficient parallel implementation of the recurrence, as detailed in \cref{app:fast_implementation}.

\textbf{The Deep \ourmethod{} Architecture.}
Following principles established in modern SSMs \citep{gu2021efficiently,orvieto2023resurrecting}, we build a hierarchy of representations by constructing a deep MP-SSM architecture composed of stacked MP-SSM blocks.
We use the output of an \model{} block as an input for the next 
one.
We visually summarize this concept at the top of Figure~\ref{fig:model}.
We note that, stacking multiple \model{} blocks allows the model to increase its effective aperture, aggregating information from further nodes. Specifically, the embedding $\Y_{k+1}$ encodes information aggregated up to the $k$-th hop. Therefore, stacking $s$  \model{} blocks, each of depth $k$, allows to aggregate information from up to $sk$ hops away.
In Appendix \ref{app:multihop}, we provide a multi-hop interpretation of a deep \ourmethod{} architecture, in the static case.




\textbf{\ourmethod{} generalizes MPNNs.}
\label{sec:equivalence} We note that our \ourmethod{} can implement its backbone MPNN, an important property that allows it to retain desired or known behavior from existing MPNNs while also generalizing it and allowing for improved information transfer, as discussed in \cref{sec:sensitivity}. 
To show that our model can implement its backbone MPNN, which in our case is based on GCN via the chosen GSO (as shown in \cref{eq:shift_operator}), we consider the static case, \ie an input sequence $[ \U_1, \ldots, \U_1 ] $, under the assumption that the MLP is a nonlinear activation $\sigma$ function. We note that this can be obtained if the weights within the MLP decoder are the identity matrices, \ie $\text{MLP}(\cdot) = \sigma(\cdot)$. Then an \model{} block with $k=1$ yields a GCN layer. In fact, if $k=1$ then \cref{eq:state_space,eq:mlp} read:
\begin{equation*}
    \X_{1} =   \U_{1} \B  \quad\Rightarrow\quad   \X_{2} =   \A \U_{1} \B \W + \U_{1}\B = \A \X_1 \W + \X_1  \quad\Rightarrow\quad
    \Y_{2} =  \sigma( \A \X_1 \W + \X_1),
\end{equation*}
which implements a GCN with a residual connection.
Then $\Y_2 $ is passed as an input to the next \model{} block, which yields a similar update rule, effectively constructing a deep GCN. 
However, we note that if $k\geq 2$, then an \model{} block deviates from the standard GCN processing.


\textbf{Final SSM heuristics.} 
\label{sec:resnet} 
If the GSO is the identity matrix $(\A=\I)$, then 
our \ourmethod{} resutls in a multi-input multi-output SSM.
This 
architecture is graph-agnostic, and it can be made resilient to vanishing and exploding gradient issues through standard deep learning heuristics such as residual connections \citep{he2016deep} and normalization layers \citep{vaswani2017attention}, with dropout being employed as a regularization technique to support the learning of robust hierarchical representations \citep{srivastava2014dropout}. In our deep \ourmethod{} architecture, we apply these heuristics between \ourmethod{} blocks, following established practices in SSMs \citep{gu2021efficiently,mamba}.
\cref{app:ablations} presents an ablation study tracing the incremental impact of each SSM heuristic on graph representation learning, progressing from a plain GCN to a deep MP-SSM architecture.
Finally, we discuss the complexity and runtimes of  \ourmethod{}  in \cref{app:complexity_runtimes}.

\section{Sensitivity Analysis}\label{sec:sensitivity}

We conduct a sensitivity analysis of \ourmethod{} via the spectral norm of the Jacobian of node features, as in \citep{topping2022understanding}. We provide an exact characterization of \ourmethod{}'s gradient flow through the graph, identify unfavourable topological structures that intensify oversquashing effects, and quantitatively assess the impact of removing nonlinearities at each recurrent step of graph diffusion, particularly in alleviating vanishing gradients in the deep regime.
\begin{remark}
\label{remark:recurrent_sensitivity}
As discussed in Section~\ref{sec:resnet}, \model{} extends graph-agnostic 
SSMs, for which established deep learning heuristics are known to effectively address vanishing/exploding gradient issues. This observation motivates our focus for sensitivity analysis on the linear recurrent equation within an MP-SSM block, as it encapsulates the core dynamics relevant to information propagation on graphs. Notably, all the other operations within a deep \model{} are independent of the graph structure. Thus, \textbf{if the linear recurrent equation supports effective information transfer, then this property naturally extends across the full MP-SSM architecture}, which is fundamentally a stack of such linear recurrences.
\end{remark}

Let $ \X_s^{(j)} $ and $\X_t^{(i)}$  denote the embeddings of nodes $j$ and $i$ at time steps $s\leq t $. We define:
\begin{definition} [Local sensitivity] \label{def:sensitivity_and_energy}
    The \emph{local sensitivity} of the features of the $i$-th node  to features of the $j$-th node,  after $t-s$ applications of message-passing aggregations, is defined as the following spectral norm:
    \begin{equation}
        \label{eq:sensitivity_local}
        \mathcal{S}_{ij}({t-s}) = \Biggl|\Biggl| \dfrac{\partial \X_t^{(i)} }{\partial \X_s^{(j)}} \Biggl|\Biggl|.
    \end{equation}
\end{definition}
\Cref{eq:sensitivity_local} measures the influence of node $ j$'s features at time $s$ on node $i$ at time $t$.

\begin{remark}
\label{remark:remark1}
If the local sensitivity between two nodes increases exponentially with $t-s$, then the learning dynamics of the MPNN are unstable; that is the typical case for linear MPNNs using the adjacency matrix without any normalization or feature normalization.
Therefore, \textbf{upper bounds on local sensitivity are linked with stable message propagation, in the deep regime}.
\end{remark}
The linearity of the recurrence of an \model{} block allows an exact computation of the Jacobian between two nodes $j,i$ at different times $s,t$, in terms of the powers of the GSO, as expressed by \Cref{eq:jacobian} in \Cref{thm:jacobiano} (for the proof, see Appendix \ref{app:proof_jacobiano}). 
\begin{theorem}[{Exact Jacobian computation in \ourmethod{}}]
    \label{thm:jacobiano}
The Jacobian of the linear recurrent equation of an \model{} block, from node $j$ at layer $s$ to node $i$ at layer $t\geq s$, can be computed exactly, and it has the following form:
\begin{equation}
    \label{eq:jacobian}
    \dfrac{\partial \X_t^{(i)} }{\partial \X_s^{(j)}} = \underbrace{(\A^{t-s})_{ij}}_{\text{scalar}} \underbrace{(\W^{\top})^{t-s}}_{\text{matrix}}.
\end{equation}
\end{theorem}
Consequently, GSOs that yield a bounded outcome under iterative multiplication promote stable \ourmethod{} dynamics, as highlighted in \Cref{remark:remark1}. 
In \Cref{lem:spectrum_gcnnorm}, we formally prove (see Appendix~\ref{app:proof_lemma}) that the GSO defined in \Cref{eq:shift_operator} exhibits this stability property, along with additional characteristics\footnote{Similar characteristics of the GSO in \Cref{eq:shift_operator} have also been discussed in \citep{oono2019graph}.} that support our theoretical analysis. 
\begin{lemma}[{Powers of symmetrically normalized adjacency with self-loops}]
    \label{lem:spectrum_gcnnorm}
Assume an undirected graph. The spectrum of the powers of the symmetric normalized adjacency matrix  $\A = \D (\tilde{\A} + \I) \D  $ is contained in the interval $[-1,1]$. The largest eigenvalue of $\A^t$ has absolute value of $1$ with corresponding eigenvector $\mathbf{d}=\text{diag}(\mathbf{D}^{\frac{1}{2}})$, for all $t\geq 1$.
In particular, the sequence of powers $ [ \A^t ]_{t\geq 1}  $ does not diverge or converge to the null matrix.
\end{lemma}
Thus, \Cref{lem:spectrum_gcnnorm} implies that the symmetrically normalized adjacency with self-loops, \ie \Cref{eq:shift_operator}, serves as a GSO that ensures stable dynamics when performing a large number of message-passing operations in the \ourmethod{}'s framework.
Moreover, for such a particular GSO, we can derive a precise approximation of the local sensitivity in the deep regime, as stated in \Cref{thm:approximation} and proved in \Cref{app:proof_approximation}.
\begin{theorem}[{Approximation deep regime}]
    \label{thm:approximation}
Assume a connected graph, and the GSO defined in \Cref{eq:shift_operator}. Then, for large values of $t-s$, the Jacobian of the linear recurrent equation of an \model{} block, from node $j$ at layer $s$ to node $i$ at layer $t\geq s$, admits the following approximation:
\begin{equation}
    \label{eq:approx_jacobian}
    \dfrac{\partial \X_t^{(i)} }{\partial \X_s^{(j)}} \approx \dfrac{\sqrt{(1+d_{i})(1+d_{j})}}{|V|+2|E|} (\W^{\top})^{t-s},
\end{equation}
where $d_l = \sum_{j=1}^{n} (\tilde{\A})_{lj}$ is the degree of the $l$-th node.
\end{theorem}
For the case of the GSO of \Cref{eq:shift_operator}, we can find a precise lower bound for the minimum local sensitivity among all possible pairs of nodes in the graph, in the deep regime (proof in \Cref{app:corollario}).
\begin{corollary}[{Lower bound minimum sensitivity}]
    \label{cor:lower_min}
Assume a connected graph, and the GSO of \Cref{eq:shift_operator}. Then, for large values of $t-s$, the following lower bound for the minimum local sensitivity of the linear recurrent equation of an \model{} block holds:
\begin{equation}
    \label{eq:lower_min}
    \dfrac{2 }{|V|+2|E|} || \W^{t-s}|| \leq \min_{i,j}\mathcal{S}_{ij}(t-s).
\end{equation}
\end{corollary}
The minimum local sensitivity is realized for pairs of nodes among which the transfer of information is the most critical due to the structure of the graph.
Therefore, \textbf{lower bounds on the minimum local sensitivity are linked to the alleviation of over-squashing}. Rewiring techniques are known to help combating this phenomenon \citep{di2023over}. \cref{cor:lower_min} proves that, without rewiring, \ourmethod{} can deal with over-squashing by increasing the norm of the recurrent weight matrix. 
In \cref{rem:bottleneck}, we construct an example of a topology that approaches the lower bound of \cref{eq:lower_min}, thus realising a worst case scenario due to over-squashing.
\begin{remark}[{Bottleneck Topologies}]\label{rem:bottleneck}
A chain of $m$ cliques of order $d$ represents a topology realising a bad scenario for \cref{eq:approx_jacobian}, since local sensitivity can reach values as low as $ \dfrac{1}{m d^2}$, scaling on long chains and large cliques, see \cref{app:wors_case} for details.
This effect is intrinsically tied to the specific topology of the graph, and it aligns with prior studies that emphasize the challenges of learning on graphs with bottleneck structures \citep{topping2022understanding}.
\end{remark}
To assess the overall gradient information flow across the entire graph in the deep regime, we define:
\begin{definition} [Global sensitivity] \label{def:global_sensitivity_and_energy}
    The \emph{global sensitivity} of node features of the overall graph after 
    $t-s$ hops of message aggregation is defined as:
    \begin{equation}
        \label{eq:sensitivity_global}
        \mathcal{S}({t-s}) = \max_{i,j} \mathcal{S}_{ij}({t-s}).
    \end{equation}
\end{definition}

\begin{remark}
\label{remark:remark1_2}
The local sensitivity between two far-apart nodes can be physiologically small due to the particular topology of the graph (e.g. bottlenecks), or it can be even 0 if two nodes are not connected by any walk. However, if the local sensitivity converges to 0, in the deep regime of large $t-s$, for all the pairs of nodes, i.e., if the global sensitivity converges to 0 regardless of the particular topology of the graph, then it means that the MPNN model is characterized by a vanishing information flow.
Therefore, \textbf{lower bounds on global sensitivity are linked to the alleviation of vanishing gradient issues, in the deep regime}. 
\end{remark}

For connected graphs, we can leverage the exact Jacobian computation of Theorem \ref{thm:jacobiano} to prove the following lower bound on the global sensitivity, see Appendix \ref{app:proof_oversquash} for the proof.
\begin{theorem}[{Lower bound global sensitivity}]
    \label{thm:oversquash}
Assume a connected graph.
The global sensitivity of the linear recurrent equation of an \model{} block
is lower bounded as follows:
\begin{equation}
\label{eq:lower_bound}
     \dfrac{\rho(\A)^{t-s}}{|V|} ||\W^{t-s} || \leq \mathcal{S}(t-s)  ,
\end{equation}
where $\rho(\A)$ is the spectral radius of the GSO. Thus, for the GSO of \cref{eq:shift_operator}, it holds the lower bound $   \dfrac{1}{|V|} ||\W^{t-s} || \leq \mathcal{S}(t-s) $.
\end{theorem}
This theoretical result demonstrates that \ourmethod{} ensures values of the global sensitivity strictly greater than zero, for any depth $t-s$ and for connected graphs with any number of nodes. 
This result cannot be guaranteed in a standard MPNN, as the nonlinearity applied at each time step increasingly contributes to vanish information as the depth increases.
We provide an extended discussion about this point in \cref{app:rationale_maximum}.
\begin{remark}
\label{remark:generality}
Note that both results of \cref{eq:jacobian,eq:lower_bound} hold for any GSO.
However, for the particular case of the symmetrically normalized adjacency with self-loops, we can provide more precise approximations and bounds.
\end{remark}

From Section \ref{sec:equivalence}, we know that \model{}  generalizes its backbone MPNNs, and the GCN architecture in particular when using \cref{eq:shift_operator} as GSO.
In \cref{thm:vanishing}, we provide an estimation of the vanishing effect caused by the application at each time step of a ReLU nonlinearity in a standard GCN compared with our \model{}, in the deep regime, as we prove in \cref{app:proof_vanishing}.
\begin{theorem}[{GCN vanishes more than \model{}}]
\label{thm:vanishing}
Let us consider a GCN 
that aggregates information from $k$ hops away, i.e., with $k$ layers, equipped with the ReLU activation function. 
Then, the GCN vanishes information at a $ 2 ^{-\frac{k}{2}}$ faster rate than our \model{} block with  $k$ linear recurrent steps. 
\end{theorem}
%
%
%

\section{Experiments}\label{sec:experiments}
We evaluate \ourmethod{} on standard benchmarks for both static and temporal graphs. For static graphs, we assess long-range propagation via synthetic shortest-path tasks (\cref{exp:gpp}) and heterophilic node classification (\cref{exp:hetero}). For temporal graphs, we test spatio-temporal forecasting performance (\cref{exp:spatiotemporal}).
Additionally, we benchmark \ourmethod{} on long-range real-world benchmarks in \cref{app:lrgb}.
\ourmethod{} is compared against state-of-the-art MPNNs, multi-hop GNNs, graph transformers, graph SSMs, and spatio-temporal models (details in \cref{app:baselines_details}). The closest baselines are MPNNs and graph SSMs. Datasets statistics and hyperparameter settings are described in \cref{app:dataset_stats,app:hyperparameters}, respectively. Code will be released upon acceptance.
We emphasize that, unlike most state-of-the-art graph models, \ourmethod{} runs at a speed comparable to that of a standard GCN (see runtime and complexity analyses in \cref{app:complexity_runtimes}), even without leveraging the optimized implementation discussed in \cref{app:fast_implementation}.

\subsection{Graph Property Prediction}\label{exp:gpp}
\begin{wraptable}{r}{8cm}
\setlength{\tabcolsep}{1pt}
\centering
\vspace{-15mm}
\footnotesize
\caption{Mean test set {\small$log_{10}(\mathrm{MSE})$}($\downarrow$) and std averaged on 4 random weight initializations on Graph Property Prediction tasks. The lower, the better. 
\one{First}, \two{second}, and \three{third} best results for each task are color-coded. 
}
\label{tab:results_gpp}
\vspace{1mm}
\footnotesize
\begin{tabular}{lccc}
\hline\toprule
\textbf{Model} &\textbf{Diameter} & \textbf{SSSP} & \textbf{Eccentricity} \\\midrule
\textbf{MPNNs} \\
$\,$ A-DGN & \two{-0.5188$_{\pm0.1812}$} & \three{-3.2417$_{\pm0.0751}$} & \two{0.4296$_{\pm0.1003}$}  \\
$\,$ DGC            & 0.6028$_{\pm0.0050}$ & -0.1483$_{\pm0.0231}$          & 0.8261$_{\pm0.0032}$\\
$\,$ GAT            & 0.8221$_{\pm0.0752}$ & 0.6951$_{\pm0.1499}$           & 0.7909$_{\pm0.0222}$  \\
$\,$ GCN            & 0.7424$_{\pm0.0466}$ & 0.9499$_{\pm0.0001}$ & 0.8468$_{\pm0.0028}$ \\
$\,$ GCNII          & 0.5287$_{\pm0.0570}$ & -1.1329$_{\pm0.0135}$          & 0.7640$_{\pm0.0355}$\\
$\,$ GIN            & 0.6131$_{\pm0.0990}$ & -0.5408$_{\pm0.4193}$          & 0.9504$_{\pm0.0007}$\\
$\,$ GRAND          & 0.6715$_{\pm0.0490}$ & -0.0942$_{\pm0.3897}$          & 0.6602$_{\pm0.1393}$ \\
$\,$ GraphCON       & 0.0964$_{\pm0.0620}$ & -1.3836$_{\pm0.0092}$ & 0.6833$_{\pm0.0074}$\\
$\,$ GraphSAGE      & 0.8645$_{\pm0.0401}$ & 0.2863$_{\pm0.1843}$           &  0.7863$_{\pm0.0207}$\\
\midrule
\textbf{Transformers} \\
$\,$ GPS & \three{-0.5121$_{\pm0.0426}$} &  \two{-3.5990$_{\pm0.1949}$}  & \three{0.6077$_{\pm0.0282}$}\\
\midrule
\textbf{Ours} \\
$\,$ \ourmethod & \one{-3.2353$_{\pm0.1735}$} & \one{-4.6321$_{\pm0.0779}$} & \one{-2.9724$_{\pm0.0271}$}\\
\bottomrule\hline      
\end{tabular}
\vspace{-7mm}
\end{wraptable}
\paragraph{Setup.} We evaluate \ourmethod{} on three synthetic tasks from \cite{gravina_adgn}: predicting graph diameter, single-source shortest paths (SSSP), and node eccentricity.
These tasks require long-range information flow, making them suitable benchmarks for evaluating propagation depth. 
We follow the original setup, data, and hyperparameters.
\vspace{-3mm}
\paragraph{Results.} 
Table~\ref{tab:results_gpp} reports results using {\small$log_{10}(\mathrm{MSE})$}. \ourmethod{} outperforms all baselines, with an average gain of 2.4 points. On the eccentricity task, it improves over A-DGN by 3.4 points, despite A-DGN being tailored for long-range tasks, and exceeds GCN (its backbone model) by over 4 points on average, despite both using the same GSO. This highlights \ourmethod{}'s superior ability to propagate information across distant nodes.

\vspace{-2mm}
\subsection{Heterophilic Benchmark}\label{exp:hetero}
\begin{table}[t]
\setlength{\tabcolsep}{2pt}
\centering
\caption{Mean test set score and std averaged over 4 random weight initializations on heterophilic 
tasks. 
\one{First}, \two{second}, and \three{third}  best results
. 
}
\label{tab:results_hetero}
\vspace{1mm}
\footnotesize
\begin{tabular}{lccccc}
\hline\toprule
\multirow{2}{*}{\textbf{Model}} & \textbf{Roman-} & \textbf{Amazon-} & \multirow{2}{*}{\textbf{Minesweep.}} & \multirow{2}{*}{\textbf{Tolokers}} & \multirow{2}{*}{\textbf{Questions}} \\
  & \textbf{empire} & \textbf{ratings} & \textbf{} & \textbf{} & \textbf{} \\
& \scriptsize{Acc $\uparrow$} & \scriptsize{Acc $\uparrow$} & \scriptsize{AUC $\uparrow$} & \scriptsize{AUC $\uparrow$} & \scriptsize{AUC $\uparrow$}\\
\midrule
\textbf{MPNNs} \\

$\,$ CO-GNN & \one{91.57$_{\pm0.32}$} &  \one{54.17$_{\pm0.37}$} &  \one{97.31$_{\pm0.41}$} &  \three{84.45$_{\pm1.17}$} & \one{80.02$_{\pm0.86}$} \\
$\,$ GAT         & 80.87$_{\pm0.30}$ & 49.09$_{\pm0.63}$ & 92.01$_{\pm0.68}$ & 83.70$_{\pm0.47}$ & 77.43$_{\pm1.20}$ \\
$\,$ Gated-GCN & 74.46$_{\pm0.54}$ & 43.00$_{\pm0.32}$ &  87.54$_{\pm1.22}$ &  77.31$_{\pm1.14}$ & 76.61$_{\pm{1.13}}$  \\
$\,$ GCN         & 73.69$_{\pm0.74}$ & 48.70$_{\pm0.63}$ & 89.75$_{\pm0.52}$ & 83.64$_{\pm0.67}$ & 76.09$_{\pm1.27}$ \\
$\,$ SAGE        & 85.74$_{\pm0.67}$ & 53.63$_{\pm0.39}$ & \three{93.51$_{\pm0.57}$} & 82.43$_{\pm0.44}$ & 76.44$_{\pm0.62}$ \\

\midrule
\multicolumn{4}{l}{\textbf{Graph Transformers}} \\
$\,$ Exphormer   & \three{89.03$_{\pm0.37}$}  &  53.51$_{\pm0.46}$  &  90.74$_{\pm0.53}$  &  83.77$_{\pm0.78}$ & 73.94$_{\pm1.06}$ \\
$\,$ GOAT        & 71.59$_{\pm1.25}$  &  44.61$_{\pm0.50}$  &  81.09$_{\pm1.02}$  &  83.11$_{\pm1.04}$ & 75.76$_{\pm1.66}$ \\
$\,$ GPS         & 82.00$_{\pm0.61}$  &  53.10$_{\pm0.42}$  &  90.63$_{\pm0.67}$  &  83.71$_{\pm0.48}$ & 71.73$_{\pm1.47}$ \\
$\,$ GT          & 86.51$_{\pm0.73}$ & 51.17$_{\pm0.66}$ & 91.85$_{\pm0.76}$ & 83.23$_{\pm0.64}$ & 77.95$_{\pm0.68}$ \\
\midrule
\multicolumn{4}{l}{\textbf{Heterophily-Designated GNNs}} \\
$\,$ FAGCN       & 65.22$_{\pm0.56}$ & 44.12$_{\pm0.30}$ & 88.17$_{\pm0.73}$ & 77.75$_{\pm1.05}$ & {77.24$_{\pm1.26}$} \\
$\,$ FSGNN       & 79.92$_{\pm0.56}$ & 52.74$_{\pm0.83}$ & 90.08$_{\pm0.70}$ & 82.76$_{\pm0.61}$ & \two{78.86$_{\pm0.92}$} \\
$\,$ GBK-GNN     & 74.57$_{\pm0.47}$ & 45.98$_{\pm0.71}$ & 90.85$_{\pm0.58}$ & 81.01$_{\pm0.67}$ & 74.47$_{\pm0.86}$ \\
$\,$ GPR-GNN     & 64.85$_{\pm0.27}$ & 44.88$_{\pm0.34}$ & 86.24$_{\pm0.61}$ & 72.94$_{\pm0.97}$ & 55.48$_{\pm0.91}$ \\
$\,$ H2GCN       & 60.11$_{\pm0.52}$ & 36.47$_{\pm0.23}$ & 89.71$_{\pm0.31}$ & 73.35$_{\pm1.01}$ & 63.59$_{\pm1.46}$ \\
\midrule
\textbf{Graph SSMs} \\
$\,$ GMN          & 87.69$_{\pm0.50}$  &  \two{54.07$_{\pm0.31}$}  &  91.01$_{\pm0.23}$  &  \two{84.52$_{\pm0.21}$} & -- \\
$\,$ GPS+Mamba  & 83.10$_{\pm0.28}$  &  45.13$_{\pm0.97}$  &  89.93$_{\pm0.54}$  &  83.70$_{\pm1.05}$ & -- \\
\midrule
\textbf{Ours} \\
$\,$ \ourmethod  & \two{90.91$_{\pm0.48}$} & \three{53.65$_{\pm0.71}$} & \two{95.33$_{\pm0.72}$} & \one{85.26$_{\pm0.93}$} &  \three{78.18$_{\pm1.34}$} \\

\bottomrule\hline      
\end{tabular}
\end{table}

\paragraph{Setup.} We evaluate \ourmethod{} on five heterophilic benchmarks from \cite{platonov2023a}: Roman-empire, Amazon-ratings, Minesweeper, Tolokers, and Questions. These tasks test the model's ability to capture complex interactions between dissimilar nodes. We follow the original data and experimental settings.
\vspace{-4mm}
\paragraph{Results.} Table~\ref{tab:results_hetero} shows that \ourmethod{} consistently performs well, achieving the highest average ranking across all tasks (see Appendix~\ref{app:complete_results}). It improves GCN by up to 17\% and surpasses transformer- and SSM-based GNNs, including methods tailored for heterophilic graphs, demonstrating strong adaptability to complex, non-homophilic structures.

\subsection{Spatio-Temporal Forecasting}\label{exp:spatiotemporal}

\begin{table}[t]
\centering
\caption{Average MSE and standard deviation ($\downarrow$) of 10 experimental repetitions. Baseline results are reported from \cite{rozemberczki2021pytorch, hmm4g, eliasof2024temporal}
\label{tab:predictive_performance}. 
\one{First}, \two{second}, and \three{third}  best methods for each task are color-coded.
}
\vspace{1mm}
\footnotesize
\setlength{\tabcolsep}{3pt}
\begin{tabular}{lcccccc}
    \hline\toprule
        \multirow{2}{*}{\textbf{Model}}
        & \textbf{Chickenpox}&  \textbf{PedalMe} & \textbf{Wikipedia} \\
        &\textbf{Hungary} & \textbf{London} & \textbf{Math}\\
\midrule
\textbf{Temporal GNNs} \\
$\,$ A3T-GCN  &    1.114$_{\pm0.008}$  &  1.469$_{\pm0.027}$     & 0.781$_{\pm0.011}$ \\
$\,$ AGCRN    &    1.120$_{\pm0.010}$  &  1.469$_{\pm0.030}$     & 0.788$_{\pm0.011}$ \\
$\,$ CDE   & \three{0.848$_{\pm0.020}$} & \three{0.810$_{\pm0.063}$} & {0.694$_{\pm0.028}$}  \\  
$\,$ DCRNN    &    1.124$_{\pm0.015}$  &  1.463$_{\pm0.019}$     & 0.679$_{\pm0.020}$ \\
$\,$ DyGrAE   &    1.120$_{\pm0.021}$  &  1.455$_{\pm0.031}$     & 0.773$_{\pm0.009}$ \\
$\,$ DynGESN & {0.907$_{\pm0.007}$} & 1.528$_{\pm0.063}$ & {0.610$_{\pm0.003}$}\\
$\,$ EGCN-O   &    1.124$_{\pm0.009}$  &  1.491$_{\pm0.024}$     & 0.750$_{\pm0.014}$ \\
$\,$ GConvGRU &    1.128$_{\pm0.011}$  &  1.622$_{\pm0.032}$     & 0.657$_{\pm0.015}$ \\
$\,$ GC-LSTM  &    1.115$_{\pm0.014}$  &  1.455$_{\pm0.023}$     & 0.779$_{\pm0.023}$ \\
$\,$ GRAND & 1.068$_{\pm0.021}$ & 1.557$_{\pm0.049}$ & 0.798$_{\pm0.034}$  \\
$\,$ GREAD & 0.983$_{\pm0.027}$ & {1.291$_{\pm0.055}$} & 0.704$_{\pm0.016}$  \\
$\,$ HMM4G & {0.939$_{\pm0.013}$} & 1.769$_{\pm0.370}$ & \two{0.542$_{\pm0.008}$}\\
$\,$ MPNN LSTM &   1.116$_{\pm0.023}$  &  1.485$_{\pm0.028}$     & 0.795$_{\pm0.010}$ \\
$\,$ TDE-GNN & \two{0.787$_{\pm0.018}$} & \two{0.714$_{\pm0.051}$} & \three{0.565$_{\pm0.017}$}\\
$\,$ T-GCN    &    1.117$_{\pm0.011}$  &  1.479$_{\pm0.012}$     & 0.764$_{\pm0.011}$ \\
\midrule
\textbf{Ours}\\
$\,$ \ourmethod & \one{0.748$_{\pm0.011}$} & \one{0.647$_{\pm0.062}$} & \one{0.509$_{\pm0.008}$} \\
\bottomrule\hline
\end{tabular}
\end{table}

\paragraph{Setup.} We evaluate \ourmethod{} 
on five forecasting datasets: Metr-LA, PeMS-Bay \citep{DCRNN}, Chickenpox Hungary, PedalMe London, and Wikipedia math \citep{rozemberczki2021pytorch}. The goal is to predict future node values from time-series data. The first two focus on traffic, while the others involve public health, delivery demand, and web activity. We follow the original settings for each dataset.

\paragraph{Results.} \ourmethod{} outperforms existing temporal GNNs across all datasets (Tables~\ref{tab:predictive_performance} and~\ref{table:appmetrLA_pemsBay}), highlighting its ability to model both spatial and temporal dependencies. These results confirm \ourmethod{}'s versatility across static and temporal graph domains.
Notably, \ourmethod{} significantly outperforms GGRNN \citep{ruiz2020gated} and GraphSSM \citep{li2024state}, see Table~\ref{table:appmetrLA_pemsBay}, two related state-space approaches for temporal graphs, thus highlighting the originality and effectiveness of our approach (see \cref{app:other_temporal_models} for an extended discussion).

\begin{table}[t]
   \setlength{\tabcolsep}{3pt}
   \centering
   \caption{Multivariate time series forecasting on the Metr-LA and PeMS-Bay datasets for Horizon 12. \one{First}, \two{second}, and \three{third}  best results for each task are color-coded. Baseline results are reported from \citep{step, staeformer, std_mae, RGDAN, AdpSTGCN}.
   \label{table:appmetrLA_pemsBay}
   }
   \vspace{1mm}
   \footnotesize
   \begin{tabular}{lccc|ccc}
      \hline\toprule
      \multirow{2}{*}{\textbf{Model}} & \multicolumn{3}{c}{Metr-LA} & \multicolumn{3}{c}{PeMS-Bay}\\
      \cmidrule{2-7}
              & MAE $\downarrow$ & RMSE $\downarrow$ & MAPE $\downarrow$ & MAE $\downarrow$ & RMSE $\downarrow$ & MAPE $\downarrow$ \\\midrule
      \multicolumn{4}{l}{\textbf{Graph Agnostic}}\\
      $\,$ HA              & 6.99  & 13.89  & 17.54\% & 3.31  & 7.54   & 7.65\%\\ 
      $\,$ FC-LSTM         & 4.37  & 8.69   & 14.00\% & 2.37  & 4.96   & 5.70\%\\ 
      $\,$ SVR             & 6.72  & 13.76  & 16.70\% & 3.28  & 7.08   & 8.00\%\\ 
      $\,$ VAR             & 6.52  & 10.11  & 15.80\% & 2.93  & 5.44   & 6.50\%\\ 
      \midrule
      \textbf{Temporal GNNs}\\
      $\,$ AdpSTGCN & 3.40 & 7.21 & \two{9.45\%} & 1.92 & 4.49 & 4.62\%\\
      $\,$ ASTGCN          & 6.51  & 12.52  & 11.64\% & 2.61  & 5.42   & 6.00\%\\  
      $\,$ DCRNN           & 3.60  & 7.60   & 10.50\% & 2.07  & 4.74   & 4.90\%\\ 
      $\,$ GMAN            & 3.44  & 7.35   & 10.07\% & {1.86}  & \three{4.32}   & {4.37\%}\\  
      $\,$ Graph WaveNet   & 3.53  & 7.37   & 10.01\% & 1.95  & 4.52   & 4.63\%\\ 
      $\,$ GTS             & 3.46  & 7.31   & 9.98\% & 1.95  & 4.43   & 4.58\%  \\  
      $\,$ MTGNN           & 3.49  & 7.23   & 9.87\% & 1.94 & 4.49 & 4.53\% \\  
      $\,$ RGDAN & \two{3.26} & \three{7.02} & 9.73\% & 1.82 & \one{4.20} & 4.28\% \\
      $\,$ STAEformer      & \three{3.34}  & \three{7.02}   & {9.70\%} & 1.88  & 4.34   & 4.41\%\\
      $\,$ STD-MAE & 3.40 & 7.07 & \three{9.59\%} & \two{1.77} & \one{4.20} & \two{4.17\%}\\
      $\,$ STEP            & {3.37}  & \two{6.99}   & {9.61\%} & \three{1.79}  & \one{4.20}   & \three{4.18\%}\\ 
      $\,$ STGCN           & 4.59  & 9.40   & 12.70\% & 2.49  & 5.69   & 5.79\%\\ 
      $\,$ STSGCN          & 5.06  & 11.66  & 12.91\% & 2.26  & 5.21   & 5.40\%\\  
    \midrule
    \textbf{Temporal Graph SSMs} \\
      $\,$ {GGRNN}          & 3.88 & 8.14 & 10.59\% & 2.34 & 5.14 & 5.21\%\\  
      $\,$ {GraphSSM-S4}          & 3.74 & 7.90 & 10.37\% & 1.98 & 4.45 & 4.77\%\\  
      \midrule
      \textbf{Ours}\\
      $\,$ \ourmethod      &  \one{3.17}     &    \one{6.86}    &  \one{9.21\%} & \one{1.62}    &    \two{4.22}    &   \one{4.05\%}      \\
      \bottomrule\hline
   \end{tabular} 
\end{table}

\section{Related Works}
\label{sec:related}

\paragraph{Learning Long-Range Dependencies on Graphs.}
While GNNs effectively model local structures via message passing, they struggle with long-range dependencies due to over-squashing and vanishing gradients \citep{alon2021on, di2023over, arroyo25vanishing}. Standard models like GCN \citep{Kipf2016}, GraphSAGE \citep{SAGE}, and GIN \citep{xu2018how} suffer from degraded performance on tasks requiring global context \citep{baek2021accurate, dwivedi2022long}, especially in heterophilic graphs \citep{luan2024heterophilicgraphlearninghandbook, snowflake}. Solutions include graph rewiring \citep{topping2022understanding, karhadkar2023fosr}, weight-space regularization \citep{gravina_adgn, gravina_swan}, and physics-inspired dynamics \citep{phdgn}. Graph Transformers (GTs) like SAN \citep{kreuzer2021rethinking}, Graphormer \citep{ying2021transformers}, and GPS \citep{graphgps} enhance expressivity using structural encodings \citep{Dwivedi2021, dwivedi2022graph}, but suffer from quadratic complexity. Scalable alternatives include sparse and linearized attention mechanisms \citep{zaheer2020big, shirzad2023exphormer,shirzad2024even, wu2023kdlgt, deng2024polynormer}, though simple MPNNs often remain competitive \citep{tonshoff2023where}.

\textbf{Learning Spatio-Temporal Interactions on Graphs.}
Temporal GNNs often combine GNNs with RNNs to model spatio-temporal dynamics \citep{gravina2023survey}. Some adopt stacked architectures that separate spatial and temporal processing \citep{GCRN, evolvegcn, mpnn_lstm, a3tgcn, cini2023scalable}, while others integrate GNNs within RNNs for joint modeling \citep{DCRNN,lrgcn, gclstm, taminglocalglobal, ruiz2020gated}. Our approach follows the latter, but goes further by embedding modern SSM principles directly into the GNN architecture, unifying spatial and temporal reasoning through linear recurrence. 
This contrasts with GGRNN \citep{ruiz2020gated}, which employs a more elaborate message-passing scheme involving nonlinear aggregation over multiple powers of the graph shift operator at each recurrent step.

\paragraph{Casting State-Space Models into Graph Learning.} 
Several recent models adopt SSMs for static graphs by imposing sequential orderings, e.g., via degree-based sorting \citep{wang2024mamba} or random walks \citep{behrouz2024graph}, often sacrificing permutation-equivariance. Spectral methods \citep{huang2024can} offer alternatives but are computationally demanding and prone to over-squashing \citep{di2023over}. 
In the temporal graph setting, GraphSSM \citep{li2024state} applies the diffusive dynamics of a GNN backbone first, followed by an SSM as a post-processing module. In contrast, our approach embeds the core principles of modern SSMs directly into the graph learning process, yielding a unified framework designed through the lens of sensitivity analysis that seamlessly supports both static and temporal graph modeling, while maintaining permutation equivariance, computational efficiency, and supporting parallel implementation.




\section{Conclusions}
In this work, we revisited Graph State-Space Models (GSSMs) through the lens of sensitivity analysis. 
While prior GSSM approaches have demonstrated empirical improvements, they typically rely on techniques that compromise core graph properties and offer only loose theoretical guarantees on information flow. 

We propose a general framework called Message-Passing State-Space Model (MP-SSM), whose formulation preserves permutation equivariance, unifies static and temporal graphs, allows for fast implementation and crucially enables \emph{exact} sensitivity analysis. This allows us to rigorously characterize node-to-node dependencies, derive precise lower bounds on vanishing gradients and over-squashing, and identify structural conditions under which information propagation is guaranteed to remain stable.

In addition to these theoretical contributions, our framework remains empirically competitive, achieving strong results across long-range, heterophilic, and spatiotemporal forecasting tasks. 
We believe this perspective positions sensitivity analysis as a principled foundation for the design of future graph state-space models.


\section*{Ethics Statement}
The research conducted in this paper conforms in every aspect with the ICLR Code of Ethics. Our study does not involve human subjects, sensitive personal data, or applications with foreseeable harmful consequences. All experiments were conducted on publicly available datasets, and no ethical concerns are anticipated regarding data usage, methodology, or findings.

\section*{Reproducibility Statement}
We provide all necessary details to implement our MP-SSM in \Cref{sec:model} and \Cref{app:fast_implementation}, and describe the setup of each experiment in \Cref{sec:experiments} and \Cref{app:exp_details}, thereby ensuring sufficient information to reproduce our results. Furthermore, all experiments are conducted on open-source datasets available online. We pledge to openly release the full codebase upon acceptance.



\bibliography{biblio}
\bibliographystyle{iclr2026_conference}

\appendix

\section{LLMs Usage}
\label{app:llm_usage}

Large Language Models (LLMs) were used as general-purpose assistive tools to improve the writing quality of this paper. Specifically, we used LLMs to help with grammar correction, rephrasing for clarity, and suggesting some improvements to the overall structure of the text. 
All LLM-generated text was carefully reviewed and edited by the authors to ensure that it accurately reflects the authors' intentions and scientific content.
No LLMs were used to generate scientific content, including but not limited to research direction, hypothesis formulation, experimental design, data analysis, or interpretation of results.

\section{Fast Parallel Implementation}
\label{app:fast_implementation}

We describe all the details to derive and implement a fast parallel implementation for the computation of an \ourmethod{} block. 

The unfolded recurrence of an \ourmethod{} block gives the following closed-form solution:
\begin{equation}
\label{eq:unfolded_temporal}
    \X_{k+1} =  \A^k  \U_{1}\B \W^k + \A^{k-1}  \U_{2}\B \W^{k-1} + \ldots +  \A  \U_{k}\B \W +  \U_{k+1}\B  .
\end{equation}
Therefore the equation of an \ourmethod{} block reads:
\begin{align}
\label{eq:state_space_unfolded}
    \X_{k+1} &= \sum_{i=0}^{k} \A^i  \U_{k+1-i}\B \W^i ,\\
\label{eq:mlp_unfolded}
    \Y_{k+1} &=  \text{MLP}(\X_{k+1}),
\end{align}

The closed-form solution of an \ourmethod{} block tells us that we could implement the whole recurrence in one shot. However, the computation of the powers of both the GSO, $\A$, and the recurrent weights, $\W$, can be extremely expensive for generic matrices and large values of $k$.
On the other hand, the powers of diagonal matrices are fairly easy to compute, since they are simply the powers of their diagonal entries.
Below, we show how to reduce a generic dense real-valued \ourmethod{} block to an equivalent diagonalised complex-valued \ourmethod{} block.

Assume the following diagonalisation of the shift operator: $ \A = \PP \mathbf{\Lambda} \PP^{-1} $.
If undirected graph, \ie $\A$ is symmetric, then by spectral theorem the $\PP$ is a real orthogonal matrix (i.e. $\PP^{-1}=\PP^\top$) and $\mathbf{\Lambda}$ is real.

Assume the following diagonalisation of the weights: $ \W = \V \mathbf{\Sigma} \V^{-1} $.
If using dense real matrices as weights, then their diagonalisation is possible only assuming complex matrices of eigenvectors $\V$ and complex eigenvalues $\mathbf{\Sigma}$. Also, note that the set of defective matrices (i.e. non-diagonalizable in $\mathbb{C}$) has zero Lebesgue measure \citep{golub2013matrix}.

Assume the following MLP equations with 2 layers: $ \text{MLP}(\X) = \phi(  \X \W_1 )  \W_2 $, where $\phi$ is a nonlinearity, and $\W_1,\W_2$ real dense matrices.\\

With the above assumptions, the MP-SSM block equations can be equivalently written as:
\begin{align}
    \X_{k+1} &= \sum_{i=0}^{k} \PP \mathbf{\Lambda}^i \PP^{-1} \U_{k+1-i}\B \V\mathbf{\Sigma}^i\V^{-1} ,\\
    \Y_{k+1} &=  \phi(  \X_{k+1} \W_1 )  \W_2 ,
\end{align}

which we can write as:
\begin{align}
    \X_{k+1} &=  \PP \biggl( \sum_{i=0}^{k} \mathbf{\Lambda}^i \PP^{-1} \U_{k+1-i}\B \V\mathbf{\Sigma}^i \biggl)\V^{-1} ,\\
    \Y_{k+1} &=  \phi(  \X_{k+1} \W_1 )  \W_2,
\end{align}

Multiply on the left side both terms by $ \PP^{-1} $ and on the right side both terms by $ \V $
\begin{equation}
    \PP^{-1} \X_{k+1} \V =   \sum_{i=0}^{k} \mathbf{\Lambda}^i \PP^{-1} \U_{k+1-i}\B \V\mathbf{\Sigma}^i
\end{equation}

If we change coordinate reference to $ \Z_{k+1}=\PP^{-1} \X_{k+1} \V $, then we can write:
\begin{align}
\label{eq:exact_recurrent}
    \Z_{k+1} &=   \sum_{i=0}^{k} \mathbf{\Lambda}^i \PP^{-1} \U_{k+1-i}\B \V\mathbf{\Sigma}^i  ,\\
\label{eq:exact_readout}
    \Y_{k+1} &= \phi( \PP \Z_{k+1} \V^{-1} \W_1 )  \W_2,
\end{align}
Equations \eqref{eq:exact_recurrent} and \eqref{eq:exact_readout} give the same exact dynamics of the equations \eqref{eq:state_space_unfolded} and \eqref{eq:mlp_unfolded}. 

The matrix of complex eigenvectors $\V$ in \eqref{eq:exact_recurrent} can be merged into the real matrix of weights $\B$ in equation \eqref{eq:recurrent}. Therefore, we can call $\hat{\B}$ a complex matrix of weights that accounts for the term $\B\V$.
Similarly, the matrix eigenvectors $\V^{-1}$ in \eqref{eq:exact_readout} can be merged into the matrix of weights $\W_1$ in equation \eqref{eq:readout}, that we call $ \hat{\W}_1 $.
To get an exact equivalence, we should exactly multiply by $\V$ and $\V^{-1}$, but merging these into learnable complex-valued matrices $ \hat{\B} $ and $ \hat{\W}_1 $ then we get similar performance.

With these new notations, we can write the equivalent diagonalised complex-valued \ourmethod{} block:
\begin{align}
\label{eq:recurrent}
    \Z_{k+1} &=   \sum_{i=0}^{k} \mathbf{\Lambda}^{i} \hat{\U}_{k+1-i}\hat{\B}\mathbf{\Sigma}^i  ,\\
\label{eq:readout}
    \Y_{k+1} &= \phi( \PP \Z_{k+1} \hat{\W}_1 )  \W_2,
\end{align}
where, in summary:
\begin{itemize}
    \item input is pre-processed as $\hat{\U}_{k+1-i} = \PP^{-1} \U_{k+1-i} $,
    \item $\bf\Lambda$ is the diagonal matrix of the eigenvalues of the GSO,
    \item learnable recurrent weights are $\hat{\B}$ (complex and dense), and $\mathbf{\Sigma}$ (complex and diagonal)
    \item learnable readout weights are $ \hat{\W}_1 $ (complex and dense), and $ \W_2 $ (real and dense) 
\end{itemize}

Equations \eqref{eq:recurrent}-\eqref{eq:readout} tell us that 
we can implement the whole recurrence efficiently in a closed-form solution that only involves powers of diagonal matrices.

\begin{figure}[h!]
    \centering
    \includegraphics[width=1.\linewidth]{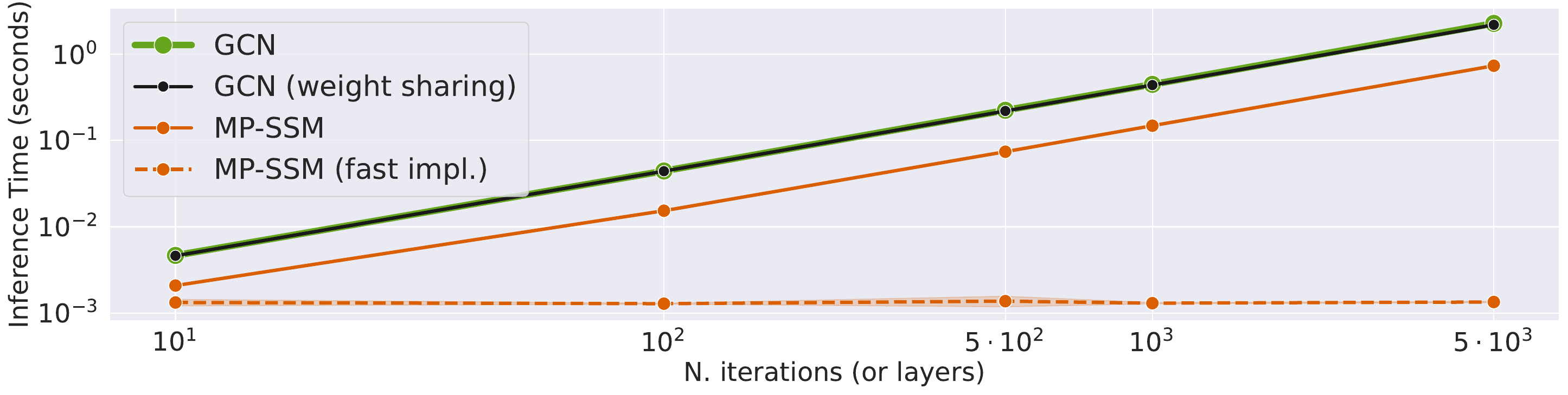}
    \caption{Inference time on a graph of $n=100$ nodes (with number of edges $3058$), input dimension $C=1$, $\text{hidden\_dim}=32$, and increasing lengths $k=10, 100, 500, 1000, 5000$. GCN is a standard GCN with $\tanh$ without residual with $k$ layers. GCN (weight sharing) is the same, but just one layer iterated $k$ times. \ourmethod{} baselines use both 1 block.}
    \label{fig:inference_time}
\end{figure}

We provide in {\cref{algo:fast}}, the pytorch-like implementation of the fast MP-SSM
, provided the input sequence $(\hat{\U}_{1}, \ldots,\hat{\U}_{k+1})$, computes in parallel the whole output sequence $(\Y_{1}, \ldots,\Y_{k+1})$.
\begin{algorithm}[H]
\caption{MP-SSM fast implementation}
\label{algo:fast}
\begin{algorithmic}[1]
\Require the input features $\texttt{x} \in \mathbb{C}^{\text{num\_steps} \times n \times C}$ (if temporal), else $\texttt{x} \in \mathbb{C}^{n \times C}$; the number of iterations (\ie k+1) \texttt{num\_steps}; the diagonal complex-valued weight matrix $\texttt{W} \in \mathbb{C}^{\text{hidden\_dim}}$; the complex-valued matrix $\texttt{B} \in \mathbb{C}^{C \times \text{hidden\_dim}}$; the eigenvalues of the GSO $\texttt{eigenvals} \in \mathbb{C}^{n}$
\Ensure $\texttt{out} \in \mathbb{C}^{\text{num\_steps} \times n \times \text{hidden\_dim}}$
\vspace{2mm}

\State $\text{powers} = \text{torch.arange}(\texttt{num\_steps})$
\vspace{1mm}
\State $\Lambda_{\text{powers}} = \text{\texttt{eigenvals}.unsqueeze}(-1).\text{pow}(\text{powers})$ \Comment{shape: $(n, \text{num\_steps})$}
\vspace{1mm}
\State $\Sigma_{\text{powers}} = \texttt{W}.\text{unsqueeze}(-1).\text{pow}(\text{powers})$ \Comment{shape: $(\text{hidden\_dim}, \text{num\_steps})$}
\vspace{1mm}
\If{\textbf{not} temporal}
    \State $\texttt{x} = \texttt{x}.\text{repeat}(\texttt{num\_steps}, 1, 1)$ \Comment{shape: $(\text{num\_steps}, n, C)$, static case}
\EndIf
\vspace{1mm}
\State $\texttt{x}_{\text{flipped}} = \text{torch.flip}(\texttt{x}, \text{dims}=[0])$ \Comment{shape: $(\text{num\_steps}, n, C)$}
\vspace{1mm}
\State $\texttt{x}_{\text{complex}} = \texttt{x}_{\text{flipped}}.\text{to}(\text{torch.cfloat})$
\vspace{1mm}
\State $\texttt{x}_B = \text{torch.matmul}(\texttt{x}_{\text{complex}}, \texttt{B})$ \Comment{shape: $(\text{num\_steps}, n, \text{hidden\_dim})$}
\vspace{1mm}
\State $\Lambda_{\text{powers}} = \Lambda_{\text{powers}}.\text{permute}(2, 0, 1)$ \Comment{shape: $(\text{num\_steps}, n, 1)$}
\vspace{1mm}
\State $\Sigma_{\text{powers}} = \Sigma_{\text{powers}}.\text{transpose}(1, 0).\text{unsqueeze}(1)$ \Comment{shape: $(\text{num\_steps}, 1, \text{hidden\_dim})$}
\vspace{1mm}
\State $\text{scaled\_x\_B} = \Lambda_{\text{powers}} \cdot \texttt{x}_B \cdot \Sigma_{\text{powers}}$
\vspace{1mm}
\State $\texttt{out} = \text{scaled\_x\_B}.\text{cumsum}(\text{dim}=0)$ \Comment{shape: $(\text{num\_steps}, n, \text{hidden\_dim})$}
\vspace{1mm}
\State $\texttt{d}_1, \texttt{d}_2, \texttt{d}_3 = \text{out.shape}$
\vspace{1mm}
\State $\texttt{x}_{\text{agg}} = \texttt{out}.\text{permute}(1, 2, 0).\text{reshape}(n, -1)$ \Comment{shape: $(n, \text{num\_steps} \cdot \text{hidden\_dim})$}

\vspace{1mm}
\State $\texttt{x}_{\text{agg}} =$ matmul(
\Statex \hspace{\algorithmicindent} $\texttt{x}=\texttt{x}_{\text{agg}},$
\Statex \hspace{\algorithmicindent} $\text{edge\_index}=\text{matrix\_p\_edge\_index},$
\Statex \hspace{\algorithmicindent} $\text{edge\_weight}=\text{matrix\_p\_edge\_weight}$
\Statex )

\vspace{1mm}
\State $\texttt{x}_{\text{agg}} = \texttt{x}_{\text{agg}}.\text{reshape}(\texttt{d}_2,\texttt{d}_3, \texttt{d}_1).\text{permute}(2, 0, 1)$
\vspace{1mm}
\State $\texttt{out} = \texttt{mlp}(\texttt{x}_{\text{agg}}, \text{batch})$
\end{algorithmic}
\end{algorithm}
We acknowledge that there is no free lunch: we achieve a one-shot parallel implementation trading off GPU memory usage, since the whole tensor of shape ($\text{num\_steps}$, $n$, $\text{hidden\_dim}$), in line 9 of \cref{algo:fast}, must fit into the GPU.
However, with sufficient GPU memory, the entire \ourmethod{} block computation occurs in $10^{-3}$ seconds, see \cref{fig:inference_time}.
As shown in \cref{fig:inference_time}, \ourmethod{} scales similarly to GCN and GCN (weight sharing), whose lines are overlapping, but it is slightly faster, owing to the lack of nonlinearity in the recurrence. This benefit grows with more iterations.
On the other hand, the fast implementation of \ourmethod{} maintains constant runtime, provided enough GPU memory.

Finally, we note that, unlike standard SSM models such as S4 and Mamba, which follow a Single-Input-Single-Output strategy (computing a separate SSM for each input channel and then mixing the results), our implementation in \cref{algo:fast} adopts a Multiple-Input-Multiple-Output strategy, enabling native handling of multivariate inputs.


\section{Proofs of the Theoretical Statements}\label{app:proofs}

Here, we provide all the proofs of lemmas, theorems, and corollaries stated in the main text. 

\subsection{Proof of Lemma \ref{lem:spectrum_gcnnorm}}\label{app:proof_lemma}

\begin{appendixlem}
Assume an undirected graph. The spectrum of the powers of the symmetric normalized adjacency matrix  $\A = \D (\tilde{\A} + \I) \D  $ is contained in the interval $[-1,1]$. The largest eigenvalue of $\A^t$ has absolute value of $1$ with corresponding eigenvector $\mathbf{d}=\text{diag}(\mathbf{D}^{\frac{1}{2}})$, for all $t\geq 1$.
In particular, the sequence of powers $ [ \A^t ]_{t\geq 1}  $ does not diverge or converge to the null matrix.
\end{appendixlem}
\begin{proof}
$\A^t = \bigl( \D (\tilde{\A} + \I) \D  \bigl) \bigl( \D (\tilde{\A} + \I) \D  \bigl) \ldots \bigl( \D (\tilde{\A} + \I) \D  \bigl) =  \D (\tilde{\A} + \I) \Bigl( \mathbf{D}^{-1} (\tilde{\A} + \I)   \Bigl)^{t-1} \D $.
Now, $\mathbf{D}^{-1} (\tilde{\A} + \I)$ is a stochastic matrix, and so also its powers are stochastic matrices.
Therefore, $ \D \A^{t} \mathbf{D}^{\frac{1}{2}} = \Bigl( \mathbf{D}^{-1} (\tilde{\A} + \I)   \Bigl)^{t} $ is a stochastic matrix.
The eigenvalues of a stochastic matrix are contained in the closed unitary disk \citep{meyer2023matrix,banerjee2016eigenvalue}. Let, $\lambda_1, \ldots, \lambda_n$ all the eigenvalues (not necessarily distinct) of such a stochastic matrix, with corresponding eigenvectors $\mathbf{v}_1, \ldots, \mathbf{v}_n$.
Thus, $ \D \A^{t} \mathbf{D}^{\frac{1}{2}} \mathbf{v}_l = \lambda_l \mathbf{v}_l $, from which it follows, multiplying both sides by $ \mathbf{D}^{\frac{1}{2}} $, that $ \A^{t} \mathbf{D}^{\frac{1}{2}} \mathbf{v}_l = \lambda_l \mathbf{D}^{\frac{1}{2}}\mathbf{v}_l $. This means that the eigenvalues of $\A^t$ are exactly the same of those of the stochastic matrix $ \D \A^{t} \mathbf{D}^{\frac{1}{2}} $ with eigenvectors $ \mathbf{D}^{\frac{1}{2}}\mathbf{v}_1, \ldots, \mathbf{D}^{\frac{1}{2}}\mathbf{v}_n$, for all $t$.
In particular, the assumption of undirected graph implies $\A$ is a symmetric matrix, thus we get that all eigenvalues of $\A^t$ are real and contained inside $[-1,1]$, for all $t$.
Since the spectral radius of a stochastic matrix is $1$, and the vector $\mathbf{1}$ with all components equal to $1$ is necessarily an eigenvector due to the row-sum being 1 for a stochastic matrix, then it follows that the largest eigenvalue of $\A^t$ is $1$ and $\mathbf{d}=\text{diag}(\mathbf{D}^{\frac{1}{2}})$ is an eigenvector corresponding to eigenvalue $1$, for all $t$.\\
To see why the sequence of powers $ [ \A^t ]_{t\geq 1}$ does not diverge or converge to the null matrix, we observe that, since $\A$ is symmetric, the Spectral Theorem implies we can diagonalize in $\mathbb{R}$ the matrix $\A = \Q \Lm \Q^{\top} $ with $\Q$ orthogonal matrix and $\Lm$ diagonal matrix of real eigenvalues. Powers of $\A$ can be written as $ \A^t = (\Q \Lm \Q^{\top} )(\Q \Lm \Q^{\top})  \ldots (\Q \Lm \Q^{\top} ) = \Q \Lm^t \Q^{\top}  $. Thus the eigenvalues of $\A^t$ are $\lambda_l^t$, for $l=1,\ldots, n$. 
We already proved that the eigenvalues $\lambda_n \leq \ldots \leq \lambda_1 $ are contained in the real interval $[-1,1]$. Hence, this ensures that the sequence of powers cannot diverge. On the other hand, we can spectrally decompose symmetric matrices as follows \citep{haykin2009neural}, $ \A^{t} = \sum_{l=1}^n \lambda_l^{t} \q_l \q_l^{\top} $, where $\q_l$ is the eigenvector corresponding to the eigenvalue $ \lambda_l $. Thus, for large values of $t $, the spectral components corresponding to eigenvalues strictly less than 1 in absolute value vanish, so the matrix $\A^t$ approaches the sum of terms corresponding to eigenvalues with absolute value equal to $1$. 
This proves that the sequence of powers cannot converge to the null matrix.
\end{proof}

\subsection{Proof of Theorem \ref{thm:jacobiano}}\label{app:proof_jacobiano}

\begin{appendixthm}
The Jacobian of the linear recurrent equation of an \model{} block, from node $j$ at layer $s$ to node $i$ at layer $t\geq s$, can be computed exactly, and it has the following form:
\begin{equation*}
    \dfrac{\partial \X_t^{(i)} }{\partial \X_s^{(j)}} = \underbrace{(\A^{t-s})_{ij}}_{\text{scalar}} \underbrace{(\W^{\top})^{t-s}}_{\text{matrix}}.
\end{equation*}
\end{appendixthm}
\begin{proof}
In this proof we use the notation $ (\M)_{ij}$ to denote the $(i,j)$ entry of a matrix $\M$, and $ \M^{(i)}$ to denote the $i$-th row of a matrix $\M$.
Let us start with the recurrent equation $ \X_{t+1} = \A \X_t \W  +  \U_{t+1} \B $. 
Therefore, the $i$-th node features are updated as follows:
$ \X_{t+1}^{(i)} = \sum_{l=1}^n (\A)_{il} \X_t^{(l)} \W + \U_{t+1}^{(i)}\B$. Now, the only term involving $\X_t^{(j)}$ is $(\A)_{ij} \X_t^{(j)} \W$. Therefore, the Jacobian reads $  \dfrac{\partial \X_{t+1}^{(i)} }{\partial \X_{t}^{(j)}  } =  \dfrac{\partial  }{\partial \X_{t}^{(j)}  } \Bigl( (\A)_{ij} \X_t^{(j)} \W \Bigl) $.
Now, given a row vector $\x \in \mathbb{R}^c$ and a square matrix $\M$, then the function $\mathbf{f}(\x) = \x \M $, whose $i$-th component is $ f_i = \sum_{l=1}^c x_l (\M)_{li} $, has derivatives $\frac{\partial f_i}{\partial \x_j} = \frac{\partial}{\partial x_j}(x_j (\M)_{ji}) = (\M)_{ji} $. Hence, the Jacobian is $\frac{\partial \mathbf{f}}{\partial \x} = \M^{\top} $.
Therefore, it holds $  \dfrac{\partial \X_{t+1}^{(i)} }{\partial \X_{t}^{(j)}  } = (\A)_{ji}\W^{\top} $.
For the case of non-consecutive time steps, we can unfold the recurrent equation $ \X_{t+1} = \A \X_t \W  +  \U_{t+1} \B $ between any two time steps $s\leq t $, as follows: 
\begin{equation}
\label{eq:folded}
 \X_{t} = \A^{t-s} \X_s \W^{t-s} + \sum_{l=0}^{t-s-1} \A^l  \U_{t-l}\B\W^i . 
\end{equation} 
From the unfolded recurrent equation \eqref{eq:folded} of a \model{} we can see that the only term involving $\X_s$ is $\A^{t-s} \X_s \W^{t-s} $. Thus, the Jacobian reads $  \dfrac{\partial \X_{t}^{(i)} }{\partial \X_{s}^{(j)}  } =  \dfrac{\partial  }{\partial \X_{s}^{(j)}  } \Bigl( ( \A^{t-s} \X_s \W^{t-s} )^{(i)} \Bigl) =  \dfrac{\partial  }{\partial \X_{s}^{(j)}  } \Bigl( (\A^{t-s})_{ij} \X_s^{(j)} \W^{t-s}  \Bigl) = (\A^{t-s})_{ij} (\W^{\top})^{t-s} $.\\
\end{proof}

\subsection{Proof of Theorem \ref{thm:approximation}}\label{app:proof_approximation}

\begin{appendixthm}
Assume a connected graph, and the GSO defined in \cref{eq:shift_operator}. Then, for large values of $t-s$, the Jacobian of the linear recurrent equation of an \model{} block, from node $j$ at layer $s$ to node $i$ at layer $t\geq s$, admits the following approximation:
\begin{equation*}
    \dfrac{\partial \X_t^{(i)} }{\partial \X_s^{(j)}} \approx \dfrac{\sqrt{(1+d_{i})(1+d_{j})}}{|V|+2|E|} (\W^{\top})^{t-s},
\end{equation*}
where $d_l = \sum_{j=1}^{n} (\tilde{\A})_{lj}$ is the degree of the $l$-th node.
\end{appendixthm}
\begin{proof}
We provide an estimation of the term $(\A^{t-s})_{ij}$ for the case of large values of $t-s$, and assuming a connected graph. We use the decomposition $ \A^{t-s} = \sum_{l=1}^n \lambda_l^{t-s} \q_l \q_l^{\top} $, where $\q_l$ is the unitary eigenvector corresponding to the eigenvalue $ \lambda_l $. As discussed in the proof of Lemma \ref{lem:spectrum_gcnnorm}, for large values of $t-s$, all the spectral components corresponding to eigenvalues strictly less than 1 (in absolute value) tend to converge to 0.
Moreover, by the Perron–Frobenius theorem for irreducible non-negative matrices \citep{horn2012matrix}, since the graph is connected and with self-loops, there is only one simple eigenvalue equal to $1$, and $-1$ cannot be an eigenvalue.
Thus it holds the approximation $ \A^{t-s} \approx  \q_1 \q_1^{\top} $. Now thanks to Lemma \ref{lem:spectrum_gcnnorm}, we know that $\mathbf{q}_1$ must be the vector $\mathbf{d}=\text{diag}(\mathbf{D}^{\frac{1}{2}})$ normalised to be unitary, and $ \mathbf{D}$ is the degree matrix of $ \tilde{\A} + \I$. Thus, $\mathbf{q_1} = \dfrac{(\sqrt{1+d_1}, \ldots, \sqrt{1+d_n})}{\sqrt{\sum_{l=1}^n (1+d_l)}}$, where $d_l = \sum_{j=1}^{n} (\tilde{\A})_{lj}$ is the degree of the $l$-th node.
Therefore, $ (\q_1 \q_1^{\top})_{ij} = \dfrac{\sqrt{(1+d_i)(1+d_j)}}{n+\sum_{l=1}^n d_l} = \dfrac{\sqrt{(1+d_i)(1+d_j)}}{|V|+2|E|}  $.
\end{proof}

\subsubsection{Example of a bad scenario for \cref{eq:approx_jacobian}}\label{app:wors_case}
\begin{figure}[h!]
    \centering
    \includegraphics[width=0.5\linewidth]{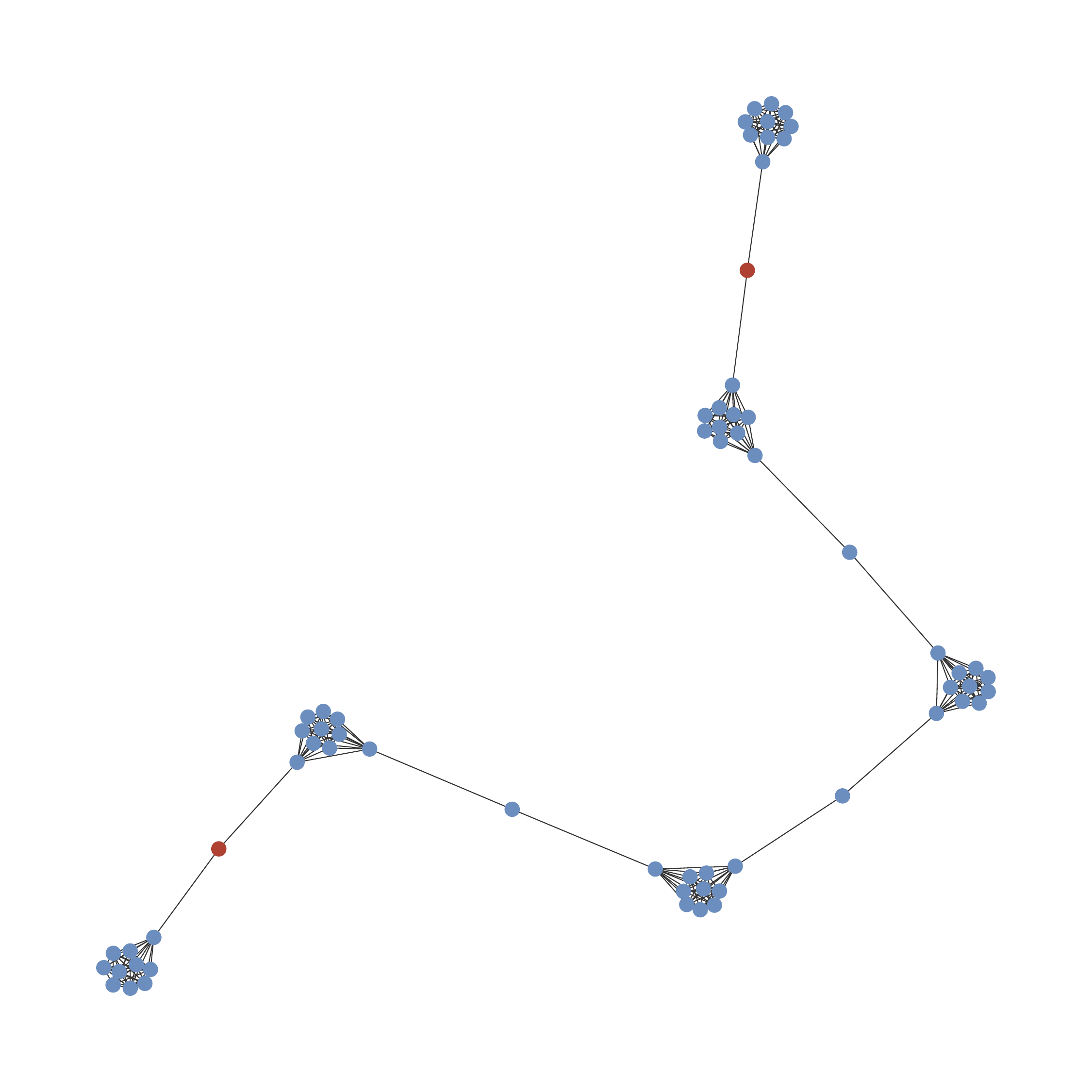}
    \caption{A chain of six cliques (containing ten nodes each) connected via bridge-nodes of degree 2. The pair of red nodes is a pair of nodes that minimizes the quantity in \cref{eq:approx_jacobian}. Note that the red nodes are $12$ hops apart, so it can be considered long-term.}
    \label{fig:worst_case_graph}
\end{figure}

\cref{fig:worst_case_graph} illustrates an example of a bad scenario for \cref{eq:approx_jacobian}, \ie a chain of $m$ cliques of order $d$ connected via bridge-nodes of degree 2 (the minimum to connect them). In the Figure, we consider $m=6$ and $d=10$.
The pair of bridge nodes $i$ and $j$ depicted in red in \cref{fig:worst_case_graph} are 12 hops apart, so it can be considered a relatively long-term interaction.

In the long-term approximation given by \cref{eq:approx_jacobian}, the local sensitivity between two bridge nodes of this topology scales as $ \frac{1}{m d^2}$, for long chains ($m$ large) and big cliques ($d$ large).
In fact, in such a graph the vast majority of nodes has degree approximately $d-1$, thus $\sum_{l=1}^n d_l \approx n(d-1) $.
Specifically, there are exactly $m-1$ nodes of degree 2 (bridge nodes), and $md$ nodes with degree approximately $d-1$. Now, $n= m-1+md \approx md$, therefore $ n+\sum_{l=1}^n d_l \approx n+n(d-1) = nd \approx md^2 $. Scaling to long chains and large cliques, this approximation becomes more accurate, and so the local sensitivity between two bridge nodes is rescaled by the term $ \frac{\sqrt{(1+d_{i})(1+d_{j})}}{n+\sum_{l=1}^n d_l}    \approx  \frac{3}{m d^2}$.

\subsection{Proof of \cref{cor:lower_min}}\label{app:corollario}

\begin{appendixcor}
Assume a connected graph, and the GSO of \cref{eq:shift_operator}. Then, for large values of $t-s$, the following lower bound for the minimum local sensitivity of the linear recurrent equation of an \model{} block holds:
\begin{equation}
    \dfrac{2 }{|V|+2|E|} || \W^{t-s}|| \leq \min_{i,j}\mathcal{S}_{ij}(t-s).
\end{equation}
\end{appendixcor}

\begin{proof}
In the deep regime, we can use the approximation of \cref{eq:approx_jacobian} of $\dfrac{\partial \X_t^{(i)} }{\partial \X_s^{(j)}} \approx \dfrac{\sqrt{(1+d_{i})(1+d_{j})}}{|V|+2|E|} (\W^{\top})^{t-s}$. Therefore, we have:
$$
\min_{i,j}\Biggl|\Biggl|\dfrac{\partial \X_t^{(i)} }{\partial \X_s^{(j)}} \Biggl|\Biggl|\approx \dfrac{1}{|V|+2|E|} \Bigl|\Bigl|(\W^{\top})^{t-s}\Bigl|\Bigl| \,\, \min_{i,j} \sqrt{(1+d_{i})(1+d_{j})} \geq \dfrac{2}{|V|+2|E|} \Bigl|\Bigl|(\W^{\top})^{t-s}\Bigl|\Bigl| ,
$$
where the last inequality holds since the minimum degree value of a node in a connected graph is $1$.
Thus, we conclude that $ \min_{i,j}\mathcal{S}_{ij}(t-s) \geq  \dfrac{2 }{|V|+2|E|} || (\W^\top)^{t-s}||= \dfrac{2 }{|V|+2|E|} || \W^{t-s}||  $, noticing that $||\W^\top||=||\W||$.
\end{proof}

\subsection{Proof of Theorem \ref{thm:oversquash}}\label{app:proof_oversquash}

\begin{appendixthm}
Assume a connected graph.
The global sensitivity of the linear recurrent equation of an \model{} block
is lower bounded as follows:
\begin{equation*}
     \mathcal{S}(t-s)  \geq  \dfrac{\rho(\A)^{t-s}}{|V|} ||\W^{t-s} || ,
\end{equation*}
where $\rho(\A)$ is the spectral radius of the GSO. Thus, for the GSO of \cref{eq:shift_operator}, it holds the lower bound $ \mathcal{S}(t-s)  \geq  \dfrac{1}{|V|} ||\W^{t-s} || $.
\end{appendixthm}
\begin{proof}
By \cref{eq:sensitivity_local,eq:sensitivity_global,eq:jacobian}
, we get $\mathcal{S}(t-s) = \max_{i,j} |(\A^{t-s})_{ij}| ||(\W^\top)^{t-s} || =  \max_{i,j} |(\A^{t-s})_{ij}| ||\W^{t-s} ||   $. 
Let us define $n=|V|$ the number of nodes.
The square of the maximum entry of an $(n,n)$ matrix $\M$ is always greater than the arithmetic mean of all the square coefficients, in other words, $ \frac{|| \M ||_F ^2 }{n^2} \leq \max_{i,j} \M_{i,j}^2 $, where $ || \M || _F $ denotes the Frobenius norm. Therefore, $ \frac{|| \M ||_F }{n} \leq \max_{i,j} |\M_{i,j}| $. 
Now, the symmetry of $\A$ implies there are $\lambda_1 , \ldots , \lambda_n $ real eigenvalues with corresponding orthonormal eigenvectors $\q_1, \ldots , \q_n$ so that we can decompose $ \A^{t-s} = \sum_{l=1}^n \lambda_l^{t-s} \q_l \q_l^{\top} $. 
Thus, the Frobenius norm is $ ||\A^{t-s}||_F =  \sqrt{\sum_{l=1}^{n} \lambda_l^{2(t-s)} 
 \,\, ||\q_l ||^2} =  \sqrt{\sum_{l=1}^{n} \lambda_l^{2(t-s)} 
 } \geq |\lambda_1|^{t-s}$, where $|\lambda_1| $ is the largest in absolute value between all the eigenvalues, i.e. the spectral radius $\rho(\A)$.
 \begin{equation}
     \max_{i,j} |(\A^{t-s})_{ij}| \geq \dfrac{||\A^{t-s}||_F}{n} \geq \dfrac{\rho(\A)^{t-s}}{n},
\end{equation}
from which we get the thesis
$$
 \mathcal{S}(t-s)  =  \max_{i,j} |(\A^{t-s})_{ij}| \, ||\W^{t-s} || \geq \dfrac{\rho(\A)^{t-s}}{n} ||\W^{t-s} ||.
$$
For the particular case of GSO given by \cref{eq:shift_operator}, the spectral radius $\rho(\A)$ is exactly 1 due to Lemma \ref{lem:spectrum_gcnnorm}.
\end{proof}

\subsection{Proof of Theorem \ref{thm:vanishing}}\label{app:proof_vanishing}

\begin{appendixthm}
Let us consider a GCN network that aggregates information from $k$ hops away, i.e., with $k$ layers, equipped with the ReLU activation function.
Then, the GCN vanishes information at a $ 2 ^{-\frac{k}{2}}$ faster rate than our \model{} block with a number $k$ of linear recurrent steps.
\end{appendixthm}
\begin{proof}
The state-update equation of a GCN with a residual connection is $ \X_{t+1} = \sigma( \A \X_t \W +\X_t)$. Therefore, the features of $i$-th node at time $t+1$ are updated as $ \X_{t+1}^{(i)} = \sigma\Bigl( \sum_{l=1}^n (\A)_{il} \X_t^{(l)} \W +  \X_t^{(i)}  \Bigl)$. Similarly to the proof of theorem \ref{thm:jacobiano}, we can write 
\begin{align*}
   \dfrac{\partial \X_{t+1}^{(i)} }{\partial \X_{t}^{(j)}  } &=  \dfrac{\partial  }{\partial \X_{t}^{(j)}  } \biggl( \sigma\Bigl( (\A)_{ij} \X_t^{(j)} \W \Bigl) \biggl) = \\
   &= \text{diag}\biggl(\sigma'\Bigl((\A)_{ij} \X_t^{(j)} \W \Bigl)\biggl) (\A)_{ij} \W^{\top} ,   
\end{align*}
where we assumed that $i \neq j$, so that the residual connection term does not appear in the derivative w.r.t. $ \X_t^{(j)} $.
Since we are considering $\sigma=\text{ReLU}$, the diagonal entries $\sigma'\Bigl((\A)_{ij} \X_t^{(j)} \W \Bigl)$ are either $0$ or $1$.
Let's assume that the components of the vector $\sigma'\Bigl((\A)_{ij} \X_t^{(j)} \W \Bigl)$ are independent and identically distributed (i.i.d.) Bernoulli random variables, each with probability $\frac{1}{2}$ of taking the value $0$.
Now, let's consider a walk $\{(i_t,j_t)\}_{t=0}^{k-1}$ of length $k$ connecting the $j$-th node at a reference time $t=0$ to the $i$-th node at time $t=k$. Then, the Jacobian of GCN along such a walk reads:
$$
\dfrac{\partial \X_{k}^{(i)} }{\partial \X_{0}^{(j)}  }
   = \prod_{t=0}^{k-1} \PP_t \M_t ,
$$
where $ \PP_t = \text{diag}\biggl(\sigma'\Bigl((\A)_{i_tj_t} \X_t^{(j_t)} \W \Bigl)\biggl)$, and $ \M_t = (\A)_{i_tj_t} \W^{\top} $. 
On the other hand, the Jacobian of the linear recurrent equation \eqref{eq:state_space} of an \model{} block, in the static case with a number $k$ of linear recurrent steps computed along the same walk reads:
$$
\dfrac{\partial \X_{k}^{(i)} }{\partial \X_{0}^{(j)}  }
   = \prod_{t=0}^{k-1}  \M_t. \\
$$
We aim to prove that, for a generic vector $\x$ with entries i.i.d. random variables distributed symmetrically about zero (e.g. according to a Normal distribution with zero mean), it holds the approximation $ || \prod_{t=0}^{k-1} \PP_t \M_t \x || \approx 2^{-\frac{k}{2}} || \prod_{t=0}^{k-1} \M_t \x || $.
We prove the thesis using a recursive argument. 
First, we observe that, denoting $ \y = \M_0 \x $, then we can write
\begin{equation}
   \label{eq:sum_bernoulli} 
   || \PP_0 \M_0 \x  ||^2 =  || \PP_0 \y  ||^2  = (p_1y_1)^2 + \ldots + (p_n y_n)^2.
\end{equation}
Now, since the $p_i$ are assumed i.i.d. Bernoulli random variables, each with probability $\frac{1}{2}$ of taking the value $0$, in the sum of \eqref{eq:sum_bernoulli}, roughly a portion of half of the contributions from $\y$ are zeroed-out due to action of $\PP_0$.
Therefore, 
\begin{equation}   
    \label{eq:approx_vanish} 
   || \PP_0 \M_0 \x  ||^2  =  || \PP_0 \y  ||^2 
   \approx \frac{1}{2} || \y ||^2 = \frac{1}{2}  || \M_0 \x ||^2  .
\end{equation}
Note that the larger the dimension of the graph $n$, the more accurate the approximation of \eqref{eq:approx_vanish}.
Therefore, we conclude that $|| \PP_0 \M_0 \x  || \approx 2^{-\frac{1}{2}}||  \M_0 \x || $. 
Now, we proceed recursively by denoting $\tilde{\x}_{t} = \PP_{t-1} \M_{t-1} \ldots \PP_0 \M_0 \x  $, and defining the scalars $ c_t = \dfrac{|| \M_t \tilde{\x}_{t} ||}{|| \tilde{\x}_{t} ||}>0$, for all $t=1, \ldots, k-1$.
Then, we can write
\begin{align*}
&|| \PP_{k-1} \M_{k-1} \PP_{k-2} \M_{k-2} \ldots \PP_0 \M_0 \x || =\\
&=|| \PP_{k-1} \M_{k-1} \tilde{\x}_{k-1} || \approx \\ &\approx 2^{-\frac{1}{2}}|| \M_{k-1}\tilde{\x}_{k-1} || =\\
&=2^{-\frac{1}{2}} c_{k-1}|| \tilde{\x}_{k-1} || = \\
&=2^{-\frac{1}{2}} c_{k-1}|| \PP_{k-2} \M_{k-2} \tilde{\x}_{k-2}  || \approx \\
&\approx 2^{-\frac{1}{2}}c_{k-1}2^{-\frac{1}{2}}c_{k-2}||   \tilde{\x}_{k-2}  || \approx \ldots \\
&\approx 2^{-\frac{k}{2}} c_{k-1}c_{k-2} \ldots c_{0} ||\x|| .
\end{align*}

On the other hand, for the case of \model{}, it reads:
\begin{align*}
&|| \M_{k-1}  \M_{k-2} \ldots  \M_0 \x || = c_{k-1} ||   \M_{k-2} \ldots  \M_0 \x || = \\
&= c_{k-1} c_{k-2}||  \M_{k-3} \ldots  \M_0 \x|| = \ldots \\
&=  c_{k-1}c_{k-2} \ldots c_{0}||\x|| .
\end{align*}
This proves that a standard GCN vanishes information $ 2 ^{-\frac{k}{2}}$ faster than \model{}.\\
We assumed weight sharing in the GCN, but the same proof holds assuming different weights $\W_1, \ldots, \W_{k}$ at each GCN layer, by simply using the same exact weight matrices for the linear equation of \ourmethod{}.
\end{proof}

\section{The Vanishing Gradient Tendency in Nonlinear MPNNs}\label{app:rationale_maximum}

Let us consider a highly connected graph without bottlenecks, such that the transfer of messages from any node to any other node is not affected by issues due to structural properties of the graph. However, in the deep regime, the presence of a nonlinearity at each time step can lead the global sensitivity (as defined in \cref{eq:sensitivity_global}) to be vanishing small. 

For an \ourmethod{} block, the local sensitivity $\mathcal{S}_{ij}(t-s)$ of the features of the $i$-th node to features of the $j$-th node after $t-s$ applications of message-passing aggregations, is exactly the norm of the Jacobian of \cref{eq:jacobian}, i.e. the norm of the product of the $(i,j)$-entry of $\A^{t-s}$ and the matrix $(\W^\top)^{t-s}$. 
For standard MPNN approaches, the local sensitivity has a more complicated expression due to nonlinearities at each aggregation step, but usually there are 3 key contributors: one from several multiplications of the shift operator (akin to $\A^{t-s}$ in our MP-SSM), one from several multiplications of the weights (akin to $(\W^\top)^{t-s}$ in our MP-SSM), and one from several multiplications of the derivative of the nonlinearity evaluated on the sequence of embeddings $\mathbf{D}(s),\mathbf{D}(s+1),..,\mathbf{D}(t)$. Usually the nonlinearity is pointwise, so $\mathbf{D}(t)$ is a diagonal matrix with entries usually in $[0,1]$, thus contributing to vanishing the gradient more and more at each time step. Hence, if the subsequent multiplications of weights and nonlinearity-based terms tend to vanish, while the powers of the shift operator $\A$ are bounded (as it is for the case of the symmetrically normalized adjacency with self-loops, proved in Lemma 4.5) then the local sensitivity tends to vanish \emph{for all pair of nodes}, for $t-s$ large enough. This will be reflected in the global sensitivity, which also will tend to vanish, for $t-s$ large enough. 
This demonstrates that global sensitivity effectively quantifies the severity of vanishing gradient issues in MPNN models plagued by this problem. Note further that the local sensitivity of the linear recurrence in each block of our MP-SSM has the exact form of $||(\A^{t-s})_{ij}(\W^\top)^{t-s}||$, while for standard MPNN approaches with nonlinearities at each time step the vanishing effect will be stronger, as we formally proved for the case of GCN in Theorem \ref{thm:vanishing}.

\section{Relation To Other Temporal Graph Models Based on State-Space Modeling}
\label{app:other_temporal_models}

In the recent literature, we can find temporal graph models that leverage the state-space approach.
The \ourmethod{} presents a simplified yet effective recurrent architecture for temporal graph modeling, offering clear advantages in architectural design when compared to alternatives such as GGRNN \citep{ruiz2020gated} or GraphSSM \citep{li2024state}.
The \ourmethod{} recurrent dynamics are governed by a simple linear diffusion on the graph:
\begin{equation}
    \label{eq:our_2}
\X_{t+1}=\A \X_t \W + \U_{t+1} \B.
\end{equation} 
In contrast, the GGRNN recurrent equation (in its simplest form, without gating mechanisms) adopts a more elaborate design:
\begin{equation}
    \label{eq:ggrnn}
    \X_{t+1} = \sigma\left(\sum_{j=0}^{K-1}\A^j \X_t \W_j+\sum_{j=0}^{K-1}\A^j \U_{t+1} \B_j\right),
\end{equation}
where multiple powers of the shift operator, $\A$, are used to aggregate information from both previous embedding $\X_t$ and current input features $\U_{t+1}$, weighted with several learnable matrices, $\W_j$ and $\B_j$, which are applied for different $j$ values, and finally, applying a nonlinearity \emph{at each time step}.

The key distinguishing feature of \ourmethod{} is the \emph{absence of nonlinearity in the recurrent update}, with the only nonlinear transformation appearing in a downstream MLP decoder, typically composed of two dense layers with an activation function in between. This feature also allows for a fast implementation of the recurrence, since it can be unfolded to get a closed-form solution, see \cref{app:fast_implementation}.
Moreover, in an \ourmethod{} block, the same weights, $\W, \B$ and MLP parameters, are shared across all time steps, ensuring \emph{strict weight sharing throughout the sequence}.
Moreover, our methodology implements a stack of \ourmethod{} blocks to build richer representations, differently from GGRNN where only one layer of recurrent computation is performed.

On the other hand, the GraphSSM model \citep{li2024state} adopts a strategy of stacking several GraphSSM blocks similar to \ourmethod{}, but their building blocks are fundamentally different from our \ourmethod{} block.
In fact, a GraphSSM block processes the spatio-temporal input sequence $[\U_t]$ in three main stages, see Appendix D.2 of \cite{li2024state}. First, a GNN backbone is applied to the input sequence, generating a corresponding sequence of node embeddings $\X_t$. Next, each embedding is mixed with the one from the previous time step $\X_{t-1}$, producing a smoothed temporal embedding $\mathbf{H}_t$. This mixed sequence $[\mathbf{H}_t]$ is then treated as a multivariate time series and passed through an SSM layer (such as S4, S5, or S6) to yield the final sequence $[\Y_t]$ as the output of a GraphSSM block.
Our approach is conceptually simpler, as it integrates both the GNN diffusive dynamics and sequence-based processing within a unified linear recurrence (\cref{eq:our_2}) followed by a shared MLP applied across time steps.
In this sense, \ourmethod{} embeds the core principles behind modern SSMs, which are the very principles that have driven the success of sequential modeling, directly into the graph processing framework. In contrast, GraphSSM merely combines GNN and SSM backbones in a modular fashion to address temporal graph tasks, without deeply integrating their underlying mechanisms.

In \cref{tab:ggrnn}, we provide a direct comparison between \ourmethod{}, GGRNN, and GraphSSM, on the Metr-LA and PeMS-Bay datasets. 
To ensure a fair and comprehensive comparison, we computed MAE, RMSE, and MAPE for all three models: \ourmethod{}, GGRNN, and GraphSSM. We used GGRNN without gating mechanisms, as it achieved the best performance on Metr-LA according to \cite[Table IV]{ruiz2020gated}, and GraphSSM-S4, since the authors reported in \cite{li2024state} that their experiments were primarily conducted using the S4 architecture. As the results show, our method consistently and significantly outperforms both GGRNN and GraphSSM across all three metrics on both datasets.
 \begin{table}[ht!]
   \setlength{\tabcolsep}{3pt}
   \centering
   \caption{Multivariate time series forecasting on the Metr-LA and PeMS-Bay datasets for Horizon 12. \textbf{Best} results for each task are in bold.
   \label{tab:ggrnn}
   }
   \vspace{1mm}
   \scriptsize
   \begin{tabular}{lcccccccc}
      \hline\toprule
      \multirow{2}{*}{\textbf{Model}} & \multicolumn{3}{c}{Metr-LA} & \multicolumn{5}{c}{PeMS-Bay}\\
      \cmidrule{2-9}
              & MAE $\downarrow$ & RMSE $\downarrow$ & MAPE $\downarrow$ & & & MAE $\downarrow$ & RMSE $\downarrow$ & MAPE $\downarrow$ \\\midrule
      $\,$ {GGRNN}          & 3.88 & 8.14 & 10.59\% & & & 2.34 & 5.14 & 5.21\%\\  
      $\,$ {GraphSSM-S4}          & 3.74 & 7.90 & 10.37\% & & & 1.98 & 4.45 & 4.77\%\\  
      \midrule
      $\,$ \ourmethod{} (ours)     &  \bf{3.17}     &    \bf{6.86}    &  \bf{9.21\%} & & & \bf{1.62}    &    \bf{4.22}    &   \bf{4.05\%}      \\
      \bottomrule\hline
   \end{tabular} 
\end{table}

\section{Multi-hop Interpretation of a Deep \ourmethod{} Architecture}\label{app:multihop}

\ourmethod{} is fundamentally different from multi-hop GNNs approaches: it operates through strictly 1-hop message passing at each iteration and does not perform aggregation from far-away hops by design. Nonetheless, to better understand its behavior in deeper architectures, we explore how a multi-hop perspective can be used for interpretation, drawing contrasts with a representative multi-hop model, Drew \citep{drew}.
For this purpose, let us consider the static case, with the input being the sequence $[\U_1, \ldots,  \U_1 ]$.
The linearity of the recurrent equation of an \ourmethod{} block allows us to unfold the recurrent equation as follows:
\begin{equation}
\label{eq:folded_mpssm}
    \X_{k+1} = \A^{k+1}\X_0\W^{k+1} + \sum_{i=0}^{k} \A^i  \U_{1}\B \W^i.
\end{equation}
Therefore, assuming a zero initial state and including the MLP into the equation, we have the following expression in the output of the first \ourmethod{} block:
\begin{equation}
\label{eq:folded_mpssm_mlp}
    \Y_{k+1} = \text{MLP}\Bigl( \sum_{i=0}^{k} \A^i  \U_{1}\B \W^i \Bigl).
\end{equation}
Due to the various powers of the shift operator $\I , \A , \A^2, \ldots, \A^k$, we can interpret \cref{eq:folded_mpssm_mlp} as a $k$-hop aggregation of the input graph $\U_1$.
Now, the sequence $[\Y_{k+1}, \ldots, \Y_{k+1}]$ is the input to the second \ourmethod{} block. Therefore, stacking the second \ourmethod{} block, and considering a residual connection from the first \ourmethod{} block, we have the following expression in the output of the second \ourmethod{} block:
\begin{equation}
\label{eq:folded_mpssm_mlp_second}
    \Y_{2(k+1)} = \Y_{k+1} + \text{MLP}\Bigl( \sum_{i=0}^{k} \A^i  \Y_{k+1}\B_2 \W_2^i \Bigl) ,
\end{equation}
where $\B_2, \W_2$, are the shared weights of the second \ourmethod{} block. 
In general, in a deep \ourmethod{} architecture of $s$ blocks, we have the following expression in the output of the $s$-th \ourmethod{} block:
\begin{equation}
\label{eq:folded_mpssm_mlp_sth}
    \Y_{s(k+1)} = \Y_{(s-1)(k+1)} + \text{MLP}\Bigl( \sum_{i=0}^{k} \A^i  \Y_{(s-1)(k+1)}\B_s \W_s^i \Bigl) .
\end{equation}
To reveal the multi-hop view, we denote $ \hat{\Y}^{(s)} = \Y_{s(k+1)} $,  $ \hat{\W}^{(s)}_{i} = \B_s \W_s^i $, and describe the deep \ourmethod{} architecture at the granularity of its blocks, as follows:
\begin{equation}
\label{eq:block_granularity}
     \hat{\Y}^{(s)} =  \hat{\Y}^{(s-1)} + \text{MLP}\Bigl( \sum_{i=0}^{k} \A^i   \hat{\Y}^{(s-1)} \hat{\W}^{(s)}_{i} \Bigl) .
\end{equation}
This multi-hop interpretation of a deep \ourmethod{} architecture resembles the DRew-GCN architecture \citep{drew}, a multi-hop MPNN employing a dynamically rewired message passing strategy with delay.
In fact, the recurrent equation of DRew-GCN, rephrased in our \ourmethod{} notation for ease of comparison, is defined as:
\begin{equation}
    \label{eq:drew}
\Y^{(s+1)}=\Y^{(s)}+\sigma\left(\sum_{i=1}^{s+1} \A(i) \Y^{(s-\tau_\nu(i))} \W^{(s)}_i\right),
\end{equation}
where $\A(i)$ is the degree-normalised shift operator that considers all the neighbors at an \emph{exact} $i$ hops from each respective root node, $\W^{(s)}_i$ are weight matrices, and $\tau_\nu(i) $ is a positive integer (the \emph{delay}) defining the temporal window for the aggregation of past embeddings.
Comparing \cref{eq:block_granularity} and \cref{eq:drew} we can summarize the following differences:
\begin{itemize}
    \item DRew aggregates information using $\A(i)$, a function of the GSO that counts neighbors at an \emph{exact} $i$ hops distance, while \ourmethod{} considers the powers of the GSO, $\A^i$, thus accounting for all the possible walks of length $i$. Similarly, the learnable weights in \ourmethod{} reflect the architectural bias induced by the recurrence, as they are structured through powers of a base matrix, specifically following the form $ \hat{\W}^{(s)}_{i} = \B_s \W_s^i $.
    \item DRew nonlinearly aggregates information via a pointwise nonlinearity $\sigma$, while \ourmethod{} employs a more expressive 2-layers MLP.
    \item \ourmethod{} uses the same features for multi-hop aggregation (corresponding to $\tau_\nu(i)\equiv 0$), whereas DRew aggregates features from previous layers with a delay $\tau_\nu(i)=\max(0,i-\nu)$, effectively introducing a temporal rewiring of the graph.
\end{itemize}
Although the unfolding of \ourmethod{} yields expressions involving powers of the GSO, this resemblance to multi-hop architectures such as DRew \citep{drew} is purely superficial. Unlike models that aggregate information from distant nodes within a single layer, \ourmethod{} performs strictly 1-hop message passing at each iteration. The higher-order GSO terms emerge naturally from the recurrence, not from an architectural bias toward multi-hop aggregation. This formulation, grounded in first principles, preserves the original graph topology and constitutes a structurally distinct approach.
We provide in \cref{tab:drew_comparison} a comparison of DRew-GCN (results taken from \cite{drew}) with our \ourmethod{} on the Peptides-func and Peptides-struct from the LRGB task \citep{dwivedi2022long}. Notably, \ourmethod{} outperforms DRew-GCN on the Peptides-struct task, suggesting that the structural architectural bias introduced by the recurrence, combined with MLP adaptivity, offers a stronger advantage than aggregating information via rewired connections from delayed past features. In contrast, on the Peptides-func task, the performance of the two models falls within each other's standard deviation, indicating no statistically significant difference between DRew-GCN, despite its dynamic rewiring strategy with delay, and \ourmethod{}.
In \cref{app:lrgb} we report an extended evaluation on the LRGB benchmark.
\begin{table}[ht!]
    \centering
    \caption{Results for Peptides-func and Peptides-struct averaged over 3 training seeds. DRew-GCN results are taken from \cite{drew}. The \textbf{best} scores are in bold.
    }\label{tab:drew_comparison}
   \vspace{1mm}
 \scriptsize
    \begin{tabular}{@{}lcc@{}}
    \hline\toprule
    \multirow{2}{*}{\textbf{Model}} & \textbf{Peptides-func}  & \textbf{Peptides-struct}        
    \\
    & \scriptsize{AP $\uparrow$} & \scriptsize{MAE $\downarrow$} 
    \\ 
    \midrule
    $\,$ DRew-GCN            & \three{69.96$_{\pm0.76}$}         & 0.2781$_{\pm0.0028}$ \\ 
    $\,$ \ourmethod{} (ours) & 69.93$_{\pm0.52}$ & \three{0.2458$_{\pm0.0017}$} \\
    \bottomrule\hline
    \end{tabular}
\end{table}

\section{Experimental Details}\label{app:exp_details}
\subsection{Employed baselines}\label{app:baselines_details}
In our experiments, the performance of our method is compared with various state-of-the-art GNN baselines from the literature. Specifically, we consider:
\begin{itemize}
    \item classical MPNN-based methods, \ie GCN~\citep{Kipf2016}, GraphSAGE~\citep{SAGE}, GAT~\citep{velivckovic2017graph}, GatedGCN~\citep{gatedgcn}, GIN~\citep{xu2018how}, {ARMA}~\citep{bianchi2021graph}, GINE~\citep{gine}, GCNII~\citep{gcnii}, and CoGNN~\citep{pmlr-v235-finkelshtein24a};
    \item heterophily-specific models, \ie H2GCN \citep{h2gcn}, CPGNN \citep{CPGNN}, FAGCN \citep{FAGCN}, GPR-GNN \citep{GPR_GNN},
FSGNN \citep{FSGNN}, GloGNN \citep{GloGNN}, GBK-GNN \citep{GBK_GNN}, and JacobiConv \citep{jacobiconv};
    \item physics-inspired MPNNs
    , \ie DGC~\citep{DGC}, GRAND~\citep{chamberlain2021grand}, GraphCON~\citep{rusch2022graph}, A-DGN~\citep{gravina_adgn}, GREAD~\citep{gread}, CDE~\citep{ijcai2023p518}, and TDE-GNN~\citep{eliasof2024temporal}; 
    \item Graph Transformers, \ie Transformer~\citep{Vaswani2017, dwivedi2021generalization}, GT~\citep{ijcai2021p0214}, SAN~\citep{san}, GPS~\citep{graphgps}, GOAT~\citep{goat}, Exphormer~\citep{shirzad2023exphormer}, NAGphormer~\citep{chen2023nagphormer}, GRIT~\citep{grit}, and GraphViT~\citep{graphvit};
    \item Higher-Order DGNs, \ie DIGL~\citep{DIGL}, MixHop~\citep{abu2019mixhop}, DRew~\citep{drew}, and GRED~\citep{ding2024recurrent}.
    \item SSM-based GNN, \ie Graph-Mamba~\citep{wang2024mamba}, GMN~\citep{behrouz2024graph}, GPS+Mamba~\citep{behrouz2024graph}, GGRNN \citep{ruiz2020gated}, and GraphSSM \citep{li2024state}.
    \item Graph-agnostic temporal predictors, \ie Historical Average (AV), SVR~\citep{svr}, and FC-LSTM~\citep{fclstm}, and VAR~\citep{var}; 
    \item Spatio-temporal GNNs, \ie DCRNN~\citep{DCRNN}, GConvGRU~\citep{GCRN}, Graph WaveNet~\citep{graphwavenet}, ASTGCN~\citep{astgcn}, STSGCN~\citep{stsgcn}, GMAN~\citep{gman}, MTGNN~\citep{mtgnn},  AGCRN~\citep{agcrn}, T-GCN~\citep{TGCN}, DyGrAE~\citep{dygrae}, EGCN-O~\citep{evolvegcn}, A3T-GCN~\citep{a3tgcn},  MPNN LSTM~\citep{mpnn_lstm}, GTS~\citep{GTS},  STEP~\citep{step}, GC-LSTM~\citep{gclstm}, DynGESN~\citep{MICHELI202285}, HMM4G~\citep{hmm4g}, STAEformer~\citep{staeformer}, RGDAN~\citep{RGDAN}, AdpSTGCN~\citep{AdpSTGCN}, and STD-MAE~\citep{std_mae}.
\end{itemize} 

\subsection{Datasets statistics}\label{app:dataset_stats}
In our experiments, we compute the performance of our \ourmethod{} on widely used benchmarks for both static and temporal graphs. Specifically, we consider:
\begin{itemize}
    \item long-range propagation tasks, \ie the three graph property prediction tasks proposed by \cite{gravina_adgn} (``Diameter'', ``SSSP'', and ``Eccentricity'') and the ``Peptide-func'' and ``Peptide-struct'' tasks from the long-range graph benchmark \citep{dwivedi2022long};
    \item heterophilic tasks, \ie ``Roman-empire'', ``Amazon-ratings'', ``Minesweeper'', ``Tolokers'', and ``Questions'' \citep{platonov2023a};
    \item temporal tasks, \ie ``Metr-LA'' and ``PeMS-Bay'' for traffic forecasting \citep{DCRNN}, and the ``Chickenpox Hungary'', ``PedalMe London'', and ``Wikipedia math'' forecasting tasks introduced by \cite{rozemberczki2021pytorch}.
\end{itemize}
In Table~\ref{tab:data_stats}, we report the statistics of the employed datasets.

\begin{table}[h]
    \centering    
    \caption{Dataset statistics}
    \label{tab:data_stats}
    \vspace{1mm}
    \footnotesize
    \begin{tabular}{clcccc}
    \hline\toprule
    &\textbf{Task}                    & \textbf{Nodes}    &  \textbf{Edges}    & \textbf{Graphs (or Timesteps)}  & \textbf{Frequency} \\\midrule
    \multirow{10}{*}{\rotatebox[origin=c]{90}{\parbox[c]{1cm}{\centering Static}}} & Diameter            & 25 - 35  &  22 - 553 & 7,040  & -- \\
    & SSSP                & 25 - 35  &  22 - 553 & 7,040  & -- \\
    & Eccentricity        & 25 - 35  &  22 - 553 & 7,040  & -- \\
    & Peptide-func        & 150.94 (avg) & 307.30 (avg)  & 15,535 & -- \\
    & Peptide-struct      & 150.94 (avg) & 307.30 (avg)  & 15,535 & -- \\	
    & Roman-empire        & 22,662   & 32,927     & 1    & -- \\
    & Amazon-ratings      & 24,492   & 93,050     & 1    & -- \\
    & Minesweeper         & 10,000   & 39,402     & 1    & -- \\
    & Tolokers            & 11,758   & 519,000    & 1    & -- \\
    & Questions           & 48,921   & 153,540    & 1    & -- \\\midrule
    \multirow{5}{*}{\rotatebox[origin=c]{90}{\parbox[c]{1cm}{\centering Temporal}}}& Metr-LA            & 207      & 1,515      & 34,272 & 5 mins\\
    & PeMS-Bay            & 325      & 2,369      & 52,116 & 5 mins\\
    & Chickenpox Hungary  & 20       & 102        & 512     & Weekly\\
    & PedalMe London      & 15       & 225        & 15      & Weekly\\
    & Wikipedia math      & 731      & 27,079     & 1,068   & Daily\\
    \bottomrule\hline
    \end{tabular}
\end{table}

\subsection{Hyperparameter space}\label{app:hyperparameters}
In Table~\ref{tab:hyperparams}, we report the grid of hyperparameters employed in our experiments by our method on all the considered benchmarks.

\begin{table}[h]
\centering
\caption{The grid of hyperparameters employed during model selection for the 
 graph property prediction tasks (\emph{GPP}), Long Range Graph Benchmark (\emph{LRGB}), heterophilic benchmarks (\emph{Hetero}), and spatio-temporal benchmarks (\emph{Temporal}).}
 \label{tab:hyperparams}
\vspace{1mm}
\scriptsize
\begin{tabular}{l|l|l|l|l}
\hline\toprule
\multirow{2}{*}{\textbf{Hyperparameters}}  & \multicolumn{4}{c}{\textbf{Values}}\\\cmidrule{2-5}
&  {\bf \emph{GPP}} & {\bf \emph{LRGB}} & {\bf \emph{Hetero}}  & {\bf \emph{Temporal}}\\\midrule
Optimizer       & Adam          & AdamW                     & AdamW & AdamW \\
Learning rate   & 0.003         &  0.001, \ 0.0005, 0.0001    & 0.001, 0.0005 ,0.0001 &  0.005, 0.001, 0.0005 ,0.0001\\ 
Weight decay    & $10^{-6}$   &  0, 0.0001, 0.001 & 0, 0.0001, 0.001 & 0, 0.0001, 0.001 \\
Dropout         & 0             & 0, 0.5 & 0, 0.4, 0.5, 0.6,& 0, 0.5 \\
N. recurrences   & 1, 5, 10, 20  & 1, 2, 4, 8, 16 & 1, 2, 4, 8, 16 & 1, 2, 4, 8, 16 \\
Embedding dim   & 10, 20, 30    & 32,64,128,256 & 32,64,128,256 & 32,64,128,256 \\
N. Blocks       & 1, 2          & 1, 2, 4, 8, 16 & 1, 2, 4, 8, 16 & 1, 2, 4, 8, 16 \\
Structure of $\mathbf{U}$ & \multicolumn{3}{c|}{$\mathbf{U} = [\U_1, \dots,\U_1]$
} & $\mathbf{U} = [\mathbf{U}_1, \mathbf{U}_2, \dots]$\\
\bottomrule\hline
\end{tabular}
\end{table}

\section{Results on the Long-Range Graph Benchmark}
\label{app:lrgb}
To further evaluate the performance of our \ourmethod{}, we consider two tasks of the Long-Range Graph Benchmark (LRGB) \citep{dwivedi2022long}.

\textbf{Setup.} We evaluate \ourmethod{} on the Peptides-func and Peptides-struct tasks from the LRGB benchmark
, which involve predicting functional and structural properties of peptides that require modeling long-range dependencies. We follow the original experimental setup and 500k parameter budget. 

\textbf{Results.} As shown in Table~\ref{tab:results_lrgb_complete}, \ourmethod{} outperforms standard MPNNs, transformer-based GNNs, and most multi-hop and SSM-based models. It achieves the highest average ranking across tasks without relying on global attention or graph rewiring. Compared to other graph SSMs, \ourmethod{} delivers strong performance while preserving permutation-equivariance.
\begin{table}[h!]

    \centering
    \caption{Results for Peptides-func and Peptides-struct averaged over 3 training seeds. Re-evaluated methods employ the 3-layer MLP readout proposed in \cite{tonshoff2023where}. Note that all MPNN-based methods include structural and positional encoding.
    The \one{first}, \two{second}, and \three{third} best scores are colored. Baseline results are reported from \cite{dwivedi2022long, drew, tonshoff2023where, graphvit, ding2024recurrent, gravina_swan}. 
    $^\ddag$ means 3-layer MLP readout and residual connections are employed.
    }\label{tab:results_lrgb_complete}
    \vspace{1mm}
\scriptsize
    \begin{tabular}{@{}lcc|c@{}}
    \hline\toprule
    \multirow{2}{*}{\textbf{Model}} & \textbf{Peptides-func}  & \textbf{Peptides-struct}    & \textbf{avg. Rank}          
    \\
    & \scriptsize{AP $\uparrow$} & \scriptsize{MAE $\downarrow$} & \scriptsize{$\downarrow$} 
    \\ \midrule  
    \textbf{MPNNs} \\
     $\,$ A-DGN      & 59.75$_{\pm0.44}$ & 0.2874$_{\pm0.0021}$ & 26.0 \\
    $\,$ GatedGCN                   & 58.64$_{\pm0.77}$ & 0.3420$_{\pm0.0013}$ & 29.0\\ 
    $\,$ GCN                        & 59.30$_{\pm0.23}$ & 0.3496$_{\pm0.0013}$ & 29.5\\ 
    $\,$ GCNII                      & 55.43$_{\pm0.78}$ & 0.3471$_{\pm0.0010}$ & 30.5\\ 
    $\,$ GINE                       & 54.98$_{\pm0.79}$ & 0.3547$_{\pm0.0045}$ & 32.0\\ 
    $\,$ GRAND      & 57.89$_{\pm0.62}$ & 0.3418$_{\pm0.0015}$ & 29.0 \\ 
    $\,$ GraphCON   & 60.22$_{\pm0.68}$ & 0.2778$_{\pm0.0018}$ & 24.0 \\ 
   $\,$ {SWAN} & 67.51$_{\pm0.39}$ & 0.2485$_{\pm0.0009}$ & 12.5\\
    \midrule
    \textbf{Multi-hop GNNs}\\
    $\,$ DIGL+MPNN           & 64.69$_{\pm0.19}$         & 0.3173$_{\pm0.0007}$ & 25.0  \\ 
    $\,$ DIGL+MPNN+LapPE     & 68.30$_{\pm0.26}$         & 0.2616$_{\pm0.0018}$ & 16.5  \\ 
    $\,$ DRew-GatedGCN       & 67.33$_{\pm0.94}$         & 0.2699$_{\pm0.0018}$  & 19.5 \\ 
    $\,$ DRew-GatedGCN+LapPE & 69.77$_{\pm0.26}$         & 0.2539$_{\pm0.0007}$  & 12.0 \\ 
    $\,$ DRew-GCN            & 69.96$_{\pm0.76}$         & 0.2781$_{\pm0.0028}$ & 14.0  \\ 
    $\,$ DRew-GCN+LapPE      & \one{71.50$_{\pm0.44}$}   & 0.2536$_{\pm0.0015}$  & 8.0 \\ 
    $\,$ DRew-GIN            & 69.40$_{\pm0.74}$         & 0.2799$_{\pm0.0016}$  & 17.5 \\ 
    $\,$ DRew-GIN+LapPE      & \two{71.26$_{\pm0.45}$}   & 0.2606$_{\pm0.0014}$  & 9.5 \\ 
    $\,$ GRED & \three{70.85$_{\pm0.27}$} & 0.2503$_{\pm0.0019}$ & 7.0  \\
    $\,$ MixHop-GCN          & 65.92$_{\pm0.36}$         & 0.2921$_{\pm0.0023}$ & 23.0  \\ 
    $\,$ MixHop-GCN+LapPE    & 68.43$_{\pm0.49}$         & 0.2614$_{\pm0.0023}$ & 15.5  \\ 
    
    \midrule
    
    \textbf{Transformers} \\
    $\,$ GraphGPS+LapPE    & 65.35$_{\pm0.41}$ & 0.2500$_{\pm0.0005}$  & 15.5 \\ 
    $\,$ Graph ViT & 69.42$_{\pm0.75}$ & \one{0.2449$_{\pm0.0016}$} & 5.5\\
    $\,$ GRIT & 69.88$_{\pm0.82}$ & \three{0.2460$_{\pm0.0012}$} & \three{5.0}\\
    $\,$ Transformer+LapPE & 63.26$_{\pm1.26}$ & 0.2529$_{\pm0.0016}$  & 19.5 \\ 
    $\,$ SAN+LapPE         & 63.84$_{\pm1.21}$ & 0.2683$_{\pm0.0043}$  & 22.0 \\ 
    \midrule
    \textbf{Modified and Re-evaluated}$^\ddag$\\
    $\,$ DRew-GCN+LapPE
    & 69.45$_{\pm0.21}$        & 0.2517$_{\pm0.0011}$ & 11.0\\
    $\,$ GatedGCN
    & 67.65$_{\pm0.47}$ & 0.2477$_{\pm0.0009}$ & 11.0\\
    $\,$ GCN
    & 68.60$_{\pm0.50}$ & \three{0.2460$_{\pm0.0007}$} & 7.5\\
    $\,$ GINE
    & 66.21$_{\pm0.67}$ & {0.2473$_{\pm0.0017}$} & 12.0\\
    $\,$ GraphGPS+LapPE
    & 65.34$_{\pm0.91}$ & 0.2509$_{\pm0.0014}$ & 17.0\\
    \midrule
    \textbf{Graph SSMs} \\
    $\,$ GMN & {70.71$_{\pm0.83}$} &  {0.2473$_{\pm0.0025}$} & \two{4.5}\\
    $\,$ Graph-Mamba & 67.39$_{\pm0.87}$ &  {0.2478$_{\pm0.0016}$} & 12.5\\
    \midrule
    \textbf{Ours}  \\
    $\,$ \ourmethod & 69.93$_{\pm0.52}$ & \two{0.2458$_{\pm0.0017}$} & \one{4.0}\\
    \bottomrule\hline
    \end{tabular}
    \end{table}

\section{Complexity and Runtimes}
\label{app:complexity_runtimes}
We discuss the theoretical complexity of our method, followed by a comparison of runtimes with other methods.

\textbf{Complexity Analysis.} Our \ourmethod{} consists of a stack of blocks. Each of them performs a linear recurrence of $k$ iterations followed by the application of a nonlinear map, as defined in \cref{eq:state_space,eq:mlp}. Note that $k$ is either the length of the temporal graph sequence or a hyperparameter. Given the similarities between the linear recurrence in \ourmethod{} and standard MPNNs, described in Section~\ref{sec:model}, the recurrence retains the complexity of standard MPNNs. Therefore, the \cref{eq:state_space} is linear in the number of node $|{V}|$ and edges $|{E}|$, achieving a time complexity of $\mathcal{O}(k\cdot(|{V}|+ |{E}|))$, with $k$ the number of iterations. Considering $\mathcal{O}(m)$ the time complexity of the MLP in \cref{eq:mlp}, then the final time complexity of one \ourmethod{} block is $\mathcal{O}(k\cdot(|{V}|+ |{E}|) + m)$ in the static case and $\mathcal{O}(k\cdot(|{V}|+ |{E}| + m))$ in the temporal case.

\textbf{Runtimes.} We provide runtimes for \ourmethod~and compare it with other methods, such as Graph GPS and GCN, in Table \ref{tab:runtimes}. In all cases, we use a model with 256 hidden dimensions and a varying depth effective by changing the number of recurrences from 2 to 16 in our \ourmethod{} with 2 \ourmethod~blocks, and the number of layers is the depth for other methods. We report the training and inference times in milliseconds, as well as the downstream performance performance obtained on the Roman-Empire dataset. As can be seen from the results in the Table, our \ourmethod{}  maintains a similar runtime to GCN, which has linear complexity with respect to the graph size, while offering strong performance at the same time. Notably, our \ourmethod{} achieves better performance than GCN and GPS, and maintains its performance as depth increases, different than GCN.
All runtimes are measured on an NVIDIA A6000 GPU with 48GB of memory.

\begin{table}[h]
\small
\centering
\caption{Training and Inference Runtime (milliseconds) and obtained node classification accuracy (\%) on the Roman-Empire dataset.}
\label{tab:runtimes}
\begin{tabular}{@{}lcccccccccc@{}}
\hline\toprule
\multirow{2}{*}{Metrics} & \multirow{2}{*}{Method} & \multicolumn{4}{c}{Depth} \\ \cmidrule(l){3-6} 
    &    & 4 & 8  & 16 & 32   \\
    \midrule
        Training (ms) &  \multirow{3}{*}{GCN} & 18.38 & 33.09 & 61.86 & 120.93\\ 
        Inference (ms) & &  9.30 & 14.64 & 27.95 &  53.55 \\
        Accuracy (\%) & & 73.60 & 61.52 & 56.86 & 52.42  \\
  \midrule
          Training (ms) & \multirow{3}{*}{GPS} & 1139.05 & 2286.96 & 4545.46 & OOM\\ 
        Inference (ms) &  & 119.10 & 208.26 & 427.89 & OOM \\
        Accuracy (\%) & & 81.97 & 81.53 & 81.88  & OOM \\
        \midrule
                  Training (ms) & \multirow{3}{*}{GPS\textsubscript{GAT+Performer} (RWSE)} & 1179.08 & 2304.77 & 4590.26 & OOM\\ 
        Inference (ms) &  & 120.11 & 209.98 & 429.03 & OOM \\
        Accuracy (\%) & & 84.89  & 87.01 & 86.94  & OOM \\
\midrule
        Training (ms) & \multirow{3}{*}{\ourmethod} &  23.19 & 41.44 & 72.09 & 141.82 \\ 
        Inference (ms) &  & 10.93 & 18.87 & 38.87 & 67.59  \\
        Accuracy (\%) & & 85.73 & 88.02 &	90.82	 & 90.91  \\
  \bottomrule\hline
\end{tabular}%
\end{table}

\section{Ablations}
\label{app:ablations}

We perform an ablation study to isolate the incremental contribution of each SSM heuristic to the performance gains in reconstructing graph-structural information that depends on learning long-range dependencies; specifically for computing quantities like the diameter of a graph, the single-source-shortest-paths (SSSP), and the eccentricity of a node, see \cref{exp:gpp} for more details on these tasks. Results of this ablation are reported in \cref{tab:ablation}.

\begin{table}[h!]
\centering
\caption{Architecture ablation study. Mean test $log_{10}(MSE)$  and std averaged on 4 random weight initialization on Graph Property Prediction tasks (\Cref{exp:gpp}). The lower, the better. The evaluation include: a nonlinear multilayer GCN (\texttt{GCN}), a linear multilayer GCN (\texttt{Linear GCN}), a linear multilayer GCN with weight sharing (\texttt{Linear GCN (ws)}), Linear GCN (ws) followed by an MLP (\texttt{1 Block Linear GCN}), a stack of multiple 1 Block Linear GCN (\texttt{Multi-Blocks Linear GCN}), and our \texttt{\ourmethod{}}, which represent a multi-blocks linear GCN with standard deep learning heuristics such as residual connections and normalisation layers between blocks. 
}
\label{tab:ablation}
\vspace{1mm}

\scriptsize
\setlength{\tabcolsep}{3pt}
\begin{tabular}{lccc}
    \hline\toprule
    \textbf{Model}
        & \textbf{Diameter} $\downarrow$ &  \textbf{SSSP} $\downarrow$& \textbf{Eccentricity} $\downarrow$\\
\midrule
GCN                         &   0.7424$_{\pm0.0466}$     & 0.9499$_{\pm0.0001}$      & 0.8468$_{\pm0.0028}$       \\
Linear GCN                  &  -2.1255$_{\pm0.0984}$     & -1.5822$_{\pm0.0002}$     & -2.1424$_{\pm0.0014}$      \\
Linear GCN (ws)             &  -2.2678$_{\pm0.1277}$     & -1.5823$_{\pm0.0001}$     & -2.1447$_{\pm0.001}$                        \\
1 Block Linear GCN          &  -2.2734$_{\pm0.1513}$     & -1.5836$_{\pm0.0025}$     & -2.1869$_{\pm0.0058}$                        \\
Multi-Blocks Linear GCN     &  -2.3531$_{\pm0.3183}$     & -1.5821$_{\pm0.0001}$     & -2.1861$_{\pm0.0066}$                        \\
\midrule
MP-SSM                      &  \textbf{-3.2353}$_{\pm0.1735}$     & \textbf{-4.6321}$_{\pm0.0779}$     & \textbf{-2.9724}$_{\pm0.0271}$ \\
\bottomrule\hline
\end{tabular}
\end{table}

The ablation conducted reveals that removing the nonlinearity from GCN yields the most significant performance improvement.
Introducing weight sharing, effectively incorporating recurrence into the linear graph diffusion process, yields a slight performance boost while considerably reducing the number of parameters. Appending an MLP at the last time step of this linear recurrent architecture does not result in statistically significant gains, except marginally for the Eccentricity task. Likewise, constructing a hierarchical block structure does not noticeably enhance performance.
These limited improvements suggest that, for the three tasks considered, the linear recurrence mechanism alone, provided a long enough recurrence, is sufficient to capture meaningful representations to reconstruct graph’s structural information. Finally, incorporating standard deep learning heuristics further strengthens the full \ourmethod{} architecture, consistently improving performance across all tasks.

\section{Extended Comparison on the Heterophilic Benchmark}\label{app:complete_results}
To further evaluate the performance of \ourmethod, we report a more complete comparison for the heterophilic task in Table~\ref{tab:results_hetero_complete}. 
Specifically, we include more MPNN-based models, graph transformers, and heterophily-designated GNNs.

In Table~\ref{tab:results_hetero_complete}, we color the top three methods. Different from the main body of the paper, here we also include sub-variants of methods in the highlighted results, providing an additional perspective on the findings. Notably, our \ourmethod{} achieves the best average ranking across all datasets in the heterophilic benchmarks.

\begin{table*}[h]
\setlength{\tabcolsep}{3pt}
\centering
\caption{Mean test set score and std averaged over 4 random weight initializations on heterophilic datasets. The higher, the better. 
\one{First}, \two{second}, and \three{third}  best results for each task are color-coded. Baseline results are reported from \cite{finkelshtein2024cooperative, behrouz2024graph, platonov2023a, muller2024attending, luan2024heterophilicgraphlearninghandbook}. ``$*$" in the rank column means that the average has been computed over less trials.
}
\label{tab:results_hetero_complete}
\vspace{1mm}
\scriptsize
\begin{tabular}{lccccc|c}
\hline\toprule
\multirow{2}{*}{\textbf{Model}} & \textbf{Roman-empire} & \textbf{Amazon-ratings} & \textbf{Minesweeper} & \textbf{Tolokers} & \textbf{Questions} & \textbf{avg. Rank}\\
& \scriptsize{Acc $\uparrow$} & \scriptsize{Acc $\uparrow$} & \scriptsize{AUC $\uparrow$} & \scriptsize{AUC $\uparrow$} & \scriptsize{AUC $\uparrow$} & \scriptsize{$\downarrow$}\\
\midrule
\multicolumn{4}{l}{\textbf{\cite{luan2024heterophilicgraphlearninghandbook}}} \\
$\,$ MLP-1 & 64.12$_{\pm0.61}$ &  38.60$_{\pm0.41}$ &  50.59$_{\pm0.83}$ & 71.89$_{\pm0.82}$ & 70.33$_{\pm0.96}$ & 41.0\\
$\,$ MLP-2 & 66.04$_{\pm0.71}$ & 49.55$_{\pm0.81}$ & 50.92$_{\pm1.25}$ & 74.58$_{\pm0.75}$ & 69.97$_{\pm1.16}$ &  34.4\\
$\,$ SGC-1 & 44.60$_{\pm0.52}$ & 40.69$_{\pm0.42}$ & 82.04$_{\pm0.77}$ & 73.80$_{\pm1.35}$ & 71.06$_{\pm0.92}$ & 38.6\\
\midrule
\multicolumn{4}{l}{\textbf{Graph-agnostic}} \\
$\,$ ResNet      & 65.88$_{\pm0.38}$ & 45.90$_{\pm0.52}$ & 50.89$_{\pm1.39}$ & 72.95$_{\pm1.06}$ & 70.34$_{\pm0.76}$ & 37.4\\
$\,$ ResNet+adj  & 52.25$_{\pm0.40}$ & 51.83$_{\pm0.57}$ & 50.42$_{\pm0.83}$ & 78.78$_{\pm1.11}$ & 75.77$_{\pm1.24}$ & 32.0\\
$\,$ ResNet+SGC  & 73.90$_{\pm0.51}$ & 50.66$_{\pm0.48}$ & 70.88$_{\pm0.90}$ & 80.70$_{\pm0.97}$ & 75.81$_{\pm0.96}$ & 29.0\\
\midrule
\textbf{MPNNs} \\
$\,$ CO-GNN($\Sigma$, $\Sigma$) & \one{91.57$_{\pm0.32}$} &  51.28$_{\pm0.56}$ &  \three{95.09$_{\pm1.18}$} &  83.36$_{\pm0.89}$ & \one{80.02$_{\pm0.86}$}  &  \three{8.0} \\ 
$\,$ CO-GNN($\mu$, $\mu$) & \two{91.37$_{\pm0.35}$} &  \one{54.17$_{\pm0.37}$} &  \one{97.31$_{\pm0.41}$} &  84.45$_{\pm1.17}$ & 76.54$_{\pm0.95}$  &  \two{6.8}\\
$\,$ GAT         & 80.87$_{\pm0.30}$ & 49.09$_{\pm0.63}$ & 92.01$_{\pm0.68}$ & 83.70$_{\pm0.47}$ & 77.43$_{\pm1.20}$ & 18.0 \\
$\,$ GAT-sep     & 88.75$_{\pm0.41}$ & 52.70$_{\pm0.62}$ & 93.91$_{\pm0.35}$ & 83.78$_{\pm0.43}$ & 76.79$_{\pm0.71}$ &  9.8 \\
$\,$ GAT (LapPE) & 84.80$_{\pm0.46}$ & 44.90$_{\pm0.73}$ & 93.50$_{\pm0.54}$ & 84.99$_{\pm0.54}$ & 76.55$_{\pm0.84}$ & 16.0 \\
$\,$ GAT (RWSE) & 86.62$_{\pm0.53}$ & 48.58$_{\pm0.41}$ & 92.53$_{\pm0.65}$ & \three{85.02$_{\pm0.67}$} & 77.83$_{\pm1.22}$ & 11.6 \\
$\,$ GAT (DEG) & 85.51$_{\pm0.56}$ & 51.65$_{\pm0.60}$ & 93.04$_{\pm0.62}$ & 84.22$_{\pm0.81}$ & 77.10$_{\pm1.23}$ & 12.6 \\
$\,$ Gated-GCN & 74.46$_{\pm0.54}$ & 43.00$_{\pm0.32}$ &  87.54$_{\pm1.22}$ &  77.31$_{\pm1.14}$ & 76.61$_{\pm{1.13}}$ &  31.4 \\
$\,$ GCN         & 73.69$_{\pm0.74}$ & 48.70$_{\pm0.63}$ & 89.75$_{\pm0.52}$ & 83.64$_{\pm0.67}$ & 76.09$_{\pm1.27}$ & 25.8 \\
$\,$ GCN (LapPE) & 83.37$_{\pm0.55}$ & 44.35$_{\pm0.36}$ & 94.26$_{\pm0.49}$ & 84.95$_{\pm0.78}$ & 77.79$_{\pm1.34}$ & 14.6 \\
$\,$ GCN (RWSE) & 84.84$_{\pm0.55}$ & 46.40$_{\pm0.55}$ & 93.84$_{\pm0.48}$ & \two{85.11$_{\pm0.77}$} & 77.81$_{\pm1.40}$ & 12.0 \\
$\,$ GCN (DEG) & 84.21$_{\pm0.47}$ & 50.01$_{\pm0.69}$ & 94.14$_{\pm0.50}$ & 82.51$_{\pm0.83}$ & 76.96$_{\pm1.21}$ & 16.4 \\
$\,$ SAGE        & 85.74$_{\pm0.67}$ & 53.63$_{\pm0.39}$ & 93.51$_{\pm0.57}$ & 82.43$_{\pm0.44}$ & 76.44$_{\pm0.62}$ & 15.6 \\
\midrule
\textbf{Graph Transformers} \\
$\,$ Exphormer   & 89.03$_{\pm0.37}$  &  53.51$_{\pm0.46}$  &  90.74$_{\pm0.53}$  &  83.77$_{\pm0.78}$ & 73.94$_{\pm1.06}$ & 16.6 \\
$\,$ NAGphormer  & 74.34$_{\pm0.77}$  &  51.26$_{\pm0.72}$  &  84.19$_{\pm0.66}$  &  78.32$_{\pm0.95}$ & 68.17$_{\pm1.53}$ & 30.6 \\
$\,$ GOAT        & 71.59$_{\pm1.25}$  &  44.61$_{\pm0.50}$  &  81.09$_{\pm1.02}$  &  83.11$_{\pm1.04}$ & 75.76$_{\pm1.66}$ & 31.2 \\
$\,$ GPS         & 82.00$_{\pm0.61}$  &  53.10$_{\pm0.42}$  &  90.63$_{\pm0.67}$  &  83.71$_{\pm0.48}$ & 71.73$_{\pm1.47}$ & 21.4 \\
$\,$ GPS\textsubscript{GCN+Performer} (LapPE) & 83.96$_{\pm0.53}$ & 48.20$_{\pm0.67}$ & 93.85$_{\pm0.41}$ & 84.72$_{\pm0.77}$ & 77.85$_{\pm1.25}$ & 12.8 \\
$\,$ GPS\textsubscript{GCN+Performer} (RWSE) & 84.72$_{\pm0.65}$ & 48.08$_{\pm0.85}$ & 92.88$_{\pm0.50}$ & 84.81$_{\pm0.86}$ & 76.45$_{\pm1.51}$ & 16.6 \\
$\,$ GPS\textsubscript{GCN+Performer} (DEG) & 83.38$_{\pm0.68}$ & 48.93$_{\pm0.47}$ & 93.60$_{\pm0.47}$ & 80.49$_{\pm0.97}$ & 74.24$_{\pm1.18}$ & 22.6 \\
$\,$ GPS\textsubscript{GAT+Performer} (LapPE) & 85.93$_{\pm0.52}$ & 48.86$_{\pm0.38}$ & 92.62$_{\pm0.79}$ & 84.62$_{\pm0.54}$ & 76.71$_{\pm0.98}$ & 14.4 \\
$\,$ GPS\textsubscript{GAT+Performer} (RWSE) & 87.04$_{\pm0.58}$ & 49.92$_{\pm0.68}$ & 91.08$_{\pm0.58}$ & 84.38$_{\pm0.91}$ & 77.14$_{\pm1.49}$ & 15.0 \\
$\,$ GPS\textsubscript{GAT+Performer} (DEG) & 85.54$_{\pm0.58}$ & 51.03$_{\pm0.60}$ & 91.52$_{\pm0.46}$ & 82.45$_{\pm0.89}$ & 76.51$_{\pm1.19}$ & 20.0 \\
$\,$ GPS\textsubscript{GCN+Transformer} (LapPE) & OOM &  OOM & 91.82$_{\pm0.41}$ & 83.51$_{\pm0.93}$ & OOM  & 33.8 \\
$\,$ GPS\textsubscript{GCN+Transformer} (RWSE) & OOM &  OOM & 91.17$_{\pm0.51}$ & 83.53$_{\pm1.06}$ & OOM  & 34.4 \\
$\,$ GPS\textsubscript{GCN+Transformer} (DEG) & OOM &  OOM & 91.76$_{\pm0.61}$ & 80.82$_{\pm0.95}$ & OOM  & 36.2 \\
$\,$ GPS\textsubscript{GAT+Transformer} (LapPE) & OOM &  OOM & 92.29$_{\pm0.61}$ & 84.70$_{\pm0.56}$ & OOM  & 30.2 \\
$\,$ GPS\textsubscript{GAT+Transformer} (RWSE) & OOM &  OOM & 90.82$_{\pm0.56}$ & 84.01$_{\pm0.96}$ & OOM  & 33.8 \\
$\,$ GPS\textsubscript{GAT+Transformer} (DEG) & OOM &  OOM & 91.58$_{\pm0.56}$ & 81.89$_{\pm0.85}$ & OOM  & 36.0 \\
$\,$ GT          & 86.51$_{\pm0.73}$ & 51.17$_{\pm0.66}$ & 91.85$_{\pm0.76}$ & 83.23$_{\pm0.64}$ & 77.95$_{\pm0.68}$ & 14.4 \\
$\,$ GT-sep      & 87.32$_{\pm0.39}$ & 52.18$_{\pm0.80}$ & 92.29$_{\pm0.47}$ & 82.52$_{\pm0.92}$ & 78.05$_{\pm0.93}$ & 12.6 \\
\midrule
\multicolumn{4}{l}{\textbf{Heterophily-Designated GNNs}} \\
$\,$ CPGNN       & 63.96$_{\pm0.62}$ & 39.79$_{\pm0.77}$ & 52.03$_{\pm5.46}$ & 73.36$_{\pm1.01}$ & 65.96$_{\pm1.95}$ & 40.0\\
$\,$ FAGCN       & 65.22$_{\pm0.56}$ & 44.12$_{\pm0.30}$ & 88.17$_{\pm0.73}$ & 77.75$_{\pm1.05}$ & {77.24$_{\pm1.26}$} & 31.0\\
$\,$ FSGNN       & 79.92$_{\pm0.56}$ & 52.74$_{\pm0.83}$ & 90.08$_{\pm0.70}$ & 82.76$_{\pm0.61}$ & \two{78.86$_{\pm0.92}$} & 18.2\\
$\,$ GBK-GNN     & 74.57$_{\pm0.47}$ & 45.98$_{\pm0.71}$ & 90.85$_{\pm0.58}$ & 81.01$_{\pm0.67}$ & 74.47$_{\pm0.86}$ & 28.0\\
$\,$ GloGNN      & 59.63$_{\pm0.69}$ & 36.89$_{\pm0.14}$ & 51.08$_{\pm1.23}$ & 73.39$_{\pm1.17}$ & 65.74$_{\pm1.19}$ & 41.0\\
$\,$ GPR-GNN     & 64.85$_{\pm0.27}$ & 44.88$_{\pm0.34}$ & 86.24$_{\pm0.61}$ & 72.94$_{\pm0.97}$ & 55.48$_{\pm0.91}$ & 38.4\\
$\,$ H2GCN       & 60.11$_{\pm0.52}$ & 36.47$_{\pm0.23}$ & 89.71$_{\pm0.31}$ & 73.35$_{\pm1.01}$ & 63.59$_{\pm1.46}$ & 39.6\\
$\,$ JacobiConv  & 71.14$_{\pm0.42}$ & 43.55$_{\pm0.48}$ & 89.66$_{\pm0.40}$ & 68.66$_{\pm0.65}$ & 73.88$_{\pm1.16}$ & 36.2\\
\midrule
\textbf{Graph SSMs} \\
$\,$ GMN          & 87.69$_{\pm0.50}$  &  \two{54.07$_{\pm0.31}$}  &  91.01$_{\pm0.23}$  &  84.52$_{\pm0.21}$ & -- & 11.0$^*$\\
$\,$ GPS + Mamba  & 83.10$_{\pm0.28}$  &  45.13$_{\pm0.97}$  &  89.93$_{\pm0.54}$  &  83.70$_{\pm1.05}$ & -- & 25.5$^*$\\
\midrule
\textbf{Ours} \\
$\,$ \ourmethod  & \three{90.91$_{\pm0.48}$} & \three{53.65$_{\pm0.71}$} & \two{95.33$_{\pm0.72}$} & \one{85.26$_{\pm0.93}$} &  \three{78.18$_{\pm1.34}$} &  \one{2.4}\\

\bottomrule\hline      
\end{tabular}
\end{table*}

\end{document}